\documentclass[runningheads,orivec]{llncs}
\usepackage[T1]{fontenc}

\usepackage{multirow}
\usepackage{adjustbox}
\usepackage{booktabs}
\usepackage{arydshln}
\usepackage{pgfplots}
\usepackage{amsmath}
\usepackage{array}
\usepackage{algorithm2e}
\usepackage{todonotes}
\usepackage[framemethod=TikZ]{mdframed}
\usetikzlibrary{patterns}

\pgfplotsset{compat=1.18}

\newcommand{\quotes}[1]{``#1''}

\renewcommand{\vec}[1]{\mathbf{#1}}

\definecolor{lightblue}{HTML}{5DA5DA}
\definecolor{lightorange}{HTML}{FAA43A}
\definecolor{lightgreen}{HTML}{60BD68}

\makeatletter
\def\adl@drawiv#1#2#3{%
        \hskip.5\tabcolsep
        \xleaders#3{#2.5\@tempdimb #1{1}#2.5\@tempdimb}%
                #2\z@ plus1fil minus1fil\relax
        \hskip.5\tabcolsep}
\newcommand{\cdashlinelr}[1]{%
  \noalign{\vskip\aboverulesep
           \global\let\@dashdrawstore\adl@draw
           \global\let\adl@draw\adl@drawiv}
  \cdashline{#1}
  \noalign{\global\let\adl@draw\@dashdrawstore
           \vskip\belowrulesep}}
\makeatother

\begin{document}

\title{Improving Generative Cross-lingual Aspect-Based Sentiment Analysis with Constrained Decoding}

\titlerunning{Generative Cross-lingual Aspect-Based Sentiment Analysis}

\author{Jakub \v{S}m\'{i}d\inst{1,2}\orcidID{0000-0002-4492-5481} \and Pavel P\v{r}ib\'{a}\v{n}\inst{1}\orcidID{0000-0002-8744-8726}  \and Pavel Kr\'{a}l\inst{1,2}\orcidID{0000-0002-3096-675X}}

\authorrunning{J. \v{S}m\'{i}d et al.}
\institute{Department of Computer Science and Engineering, University of West Bohemia in Pilsen, Univerzitn\'{i}, Pilsen, Czech Republic
\and NTIS - New Technologies for the Information Society, University of West Bohemia in Pilsen, Univerzitn\'{i}, Pilsen, Czech Republic \\
\url{https://nlp.kiv.zcu.cz/}
\\
\email{\{jaksmid,pribanp,pkral\}@kiv.zcu.cz}}

\maketitle

\begin{abstract}
While aspect-based sentiment analysis (ABSA) has made substantial progress, challenges remain for low-resource languages, which are often overlooked in favour of English. Current cross-lingual ABSA approaches focus on limited, less complex tasks and often rely on external translation tools. This paper introduces a novel approach using constrained decoding with sequence-to-sequence models, eliminating the need for unreliable translation tools and improving cross-lingual performance by 5\% on average for the most complex task. The proposed method also supports multi-tasking, which enables solving multiple ABSA tasks with a single model, with constrained decoding boosting results by more than 10\%. 

We evaluate our approach across seven languages and six ABSA tasks, surpassing state-of-the-art methods and setting new benchmarks for previously unexplored tasks. Additionally, we assess large language models (LLMs) in zero-shot, few-shot, and fine-tuning scenarios. While LLMs perform poorly in zero-shot and few-shot settings, fine-tuning achieves competitive results compared to smaller multilingual models, albeit at the cost of longer training and inference times.

We provide practical recommendations for real-world applications, enhancing the understanding of cross-lingual ABSA methodologies. This study offers valuable insights into the strengths and limitations of cross-lingual ABSA approaches, advancing the state-of-the-art in this challenging research domain.
\end{abstract}

\section{Introduction}
\label{sec:intro}
Aspect-based sentiment analysis (ABSA) is a natural language processing (NLP) task that focuses on identifying sentiment associated with specific aspects or features of a product or service, providing a more detailed examination than traditional sentiment analysis. ABSA finds practical applications in diverse fields such as product marketing, customer feedback analysis, and reputation management. However, despite its significance, ABSA research has predominantly concentrated on English, leaving a gap in understanding the challenges of conducting ABSA in other languages, notably due to the lack of annotated data. Nevertheless, manual data annotation is resource-intensive and expensive, especially for languages with smaller speaker populations. To address this challenge, researchers have turned to cross-lingual sentiment analysis as a promising solution. This approach involves transferring knowledge from a \textit{source language}, typically a resource-rich language with a large amount of annotated data, to a \textit{target language}, which usually has limited resources, enabling the model to leverage the information of annotated data from the source language to perform sentiment analysis in the target language.

ABSA involves several sentiment elements~\cite{SMID2025103073}: 1) aspect term ($a$) is a word or phrase that represents the aspect within the text, 2) aspect category ($c$) defines unique aspects of an entity, and 3) sentiment polarity ($p$) indicates the orientation of the sentiment. For example, in a sentence \textit{\quotes{Delicious tea}}, these elements are \textit{\quotes{tea}}, \textit{\quotes{drinks quality}}, \textit{\quotes{Delicious}}, and \textit{\quotes{positive}}. ABSA encompasses various tasks, including simple ones that focus on predicting a single sentiment element, such as aspect term extraction (ATE)~\cite{pontiki-etal-2014-semeval} or aspect category detection (ACD)~\cite{pontiki-etal-2014-semeval}. Recently, there has been a shift towards compound ABSA tasks that predict multiple sentiment elements simultaneously, making them more challenging as the number of elements to predict increases. These tasks include aspect category term extraction (ACTE)~\cite{pontiki-etal-2015-semeval}, aspect category sentiment analysis (ACSA)~\cite{schmitt-etal-2018-joint}, end-to-end ABSA (E2E-ABSA)~\cite{wang2018towards}, and target-aspect-sentiment detection (TASD)~\cite{wan2020target}. Table~\ref{tab:absa-tasks} shows the output format of selected ABSA tasks.

\begin{table}[ht!]
    \centering
    \caption{Output format for selected ABSA tasks for input: \textit{\quotes{Delicious tea but pricey soup}}. Symbols $a$, $c$, and $p$ denote aspect term, aspect category, and sentiment polarity, respectively.}
    \begin{adjustbox}{width=0.65\linewidth}
        \begin{tabular}{@{}llll@{}}
            \toprule
            \textbf{Type}             &\textbf{Task} &  \textbf{Output}     & \textbf{Example output}  \\                        \midrule
            \multirow{2}{*}{Simple} & ATE         &  \{$a$\}               &  \{\quotes{tea}, \quotes{soup}\} \\
            &ACD         &  \{$c$\}               &  \{drinks, food\} \\ \cdashlinelr{1-4}
            \multirow{4}{*}{Compound} & E2E-ABSA      &  \{($a$, $p$)\}      & \{(\quotes{tea}, POS), (\quotes{soup}, NEG)\}               \\
            &ACSA      &  \{($c$, $p$)\}      & \{(drinks, POS), (food, NEG)\}               \\
            &ACTE          &  \{($a$, $c$)\}      & \{(\quotes{tea}, drinks), (\quotes{soup}, food)\}           \\
            &TASD          & \{($a$, $c$, $p$)\} & \{(\quotes{tea}, drinks, POS), (\quotes{soup}, food, NEG)\} \\
            \bottomrule
        \end{tabular}
    \end{adjustbox}
	\label{tab:absa-tasks}
\end{table}

Multilingual pre-trained language models (mPLMs) based on the Transformer architecture~\cite{vaswani2017attention}, such as mBERT~\cite{devlin-etal-2019-bert}, XLM-R~\cite{conneau-etal-2020-unsupervised}, and mT5~\cite{xue-etal-2021-mt5}, have become the standard for cross-lingual transfer in NLP~\cite{pmlr-v119-hu20b}. Typically, these multilingual models are fine-tuned on labelled data in the source language and applied directly to target language data for inference, leveraging the language knowledge acquired during pre-training. The zero-shot method uses only data from the source language for fine-tuning and relies solely on pre-training for language-specific knowledge, which may not adequately cover low-resource languages. Using translated target language data with projected labels presents a potential solution, although its effectiveness depends on the quality of translation and label projection. However, this approach is also relatively expensive and complex.

Large language models (LLMs), such as ChatGPT~\cite{chatgpt-2022} and LLaMA-based models~\cite{touvron2023llama2}, excel in zero-shot and few-shot scenarios across various NLP tasks~\cite{bang2023multitask}. LLMs typically comprise billions of parameters, with models exceeding 10 billion parameters considered large~\cite{zhang2023sentiment,minaee2024large}\footnote{This work considers models with 7 billion parameters and more as large models.}. Their size makes traditional fine-tuning challenging, making \textit{prompting} a preferred alternative, where task descriptions guide model outputs without extensive fine-tuning. However, fine-tuning approaches on adequate data consistently outperform LLMs on compound ABSA tasks~\cite{zhang2023sentiment,gou-etal-2023-mvp}. Techniques like QLoRA~\cite{dettmers2023qlora} offer a way to fine-tune large models efficiently on a single GPU, yet fine-tuned open-source LLMs for cross-lingual ABSA are still underexplored~\cite{SMID2025103073}.

Existing research on cross-lingual ABSA presents several key limitations. First, there is a considerably lower amount of cross-lingual ABSA research compared to monolingual ABSA. Second, the scope of cross-lingual ABSA tasks is narrow, particularly in compound tasks where multiple sentiment elements are involved. While recent English ABSA research has increasingly focused on these complex compound tasks, cross-lingual studies often address only simpler tasks and explores only one compound task, leaving most of the compound tasks underexplored. Third, most of the current cross-lingual ABSA research relies on older machine learning techniques, and there is a scarcity of studies employing modern Transformer-based models, with the exception of some works that use encoder-only Transformer models. In contrast, state-of-the-art performance in monolingual ABSA is frequently achieved with sequence-to-sequence approaches, which remain unexplored in cross-lingual settings. Fourth, most existing cross-lingual ABSA approaches depend heavily on external translation tools. Although these tools help bridge the language gap, they introduce additional complexity and potential errors, which can negatively affect the quality of the results. Finally, there has been little investigation into the potential of fine-tuned open-source LLMs for cross-lingual ABSA. While preliminary research~\cite{smid-etal-2024-llama} in monolingual settings suggests that LLMs can deliver state-of-the-art results for ABSA tasks in English, their cross-lingual applications remain largely unexplored.

The main motivation of this paper is to address the limited research on compound cross-lingual ABSA tasks and the absence of sequence-to-sequence approaches, which are critical in monolingual ABSA, while also reducing the reliance on external translation tools that introduce complexity and potential inaccuracies. Another motivation is the lack of research on cross-lingual capabilities of LLMs for ABSA. To address these issues, we introduce a novel sequence-to-sequence method enhanced with constrained decoding. This technique refines a model’s token generation process to ensure that predictions adhere to the necessary output structure by restricting them to acceptable tokens. Our approach eliminates the dependency on external translation tools, which can be unreliable and may introduce errors in cross-lingual ABSA tasks. Additionally, we evaluate several LLMs in cross-lingual settings and compare them to smaller multilingual models.

We achieve excellent results in zero-shot cross-lingual settings and evaluate our approach on seven languages and across six tasks, four of which are compound. Furthermore, given the limited research on LLMs for both monolingual and cross-lingual ABSA, we explore their capabilities for handling compound ABSA tasks, assessing their performance and potential in this context. Our main contributions are as follows:
\begin{itemize}
    \item We propose a novel method that employs constrained decoding combined with sequence-to-sequence models. Constrained decoding significantly improves cross-lingual ABSA compared to the methods that do not utilize this enhancement. 
    The proposed approach outperforms significantly state-of-the-art results for existing cross-lingual ABSA tasks and eliminates the need for external translation tools. To the best of our knowledge, this study represents the first application of sequence-to-sequence models and large language models for cross-lingual ABSA.
    \item We conduct extensive experiments on benchmark datasets across seven languages, evaluating performance on six distinct ABSA tasks. In addition to improving results on established cross-lingual and monolingual ABSA tasks, we pioneer the evaluation of several previously unexplored cross-lingual ABSA tasks.
    \item Our methodology supports multi-tasking capabilities, enabling the simultaneous solution of multiple ABSA tasks using a single model. Constrained decoding proves particularly effective in multi-tasking settings, consistently enhancing results by over 10\%.
    \item We systematically evaluate the performance of various LLMs in zero-shot, few-shot, and fine-tuning scenarios across monolingual and cross-lingual settings. Our findings demonstrate that fine-tuning LLMs is necessary to achieve results comparable to those of smaller multilingual models. Additionally, we show that more advanced LLMs significantly outperform their counterparts, emphasizing the importance of model selection. This evaluation highlights the strengths and limitations of each approach, providing valuable insights for selecting appropriate models in different ABSA applications.
    \item We include a comparison of training and inference time requirements for different models. This analysis reveals that while some models achieve high performance, they come with substantial computational costs, highlighting the efficiency advantages of our method.
    \item Based on our extensive experimentation and analysis, we propose practical recommendations tailored for diverse scenarios in both cross-lingual and monolingual ABSA. These recommendations aim to guide researchers and practitioners towards effective model selection and deployment strategies in real-world applications.
\end{itemize}

This paper extends our previous publication~\cite{smid-etal-icaart-25} and together, the two works pioneer the use of sequence-to-sequence models and LLMs for cross-lingual ABSA. Our approach differs from much of the existing research, which often depends on external translation tools and yields mixed results. Moreover, we explore cross-lingual ABSA in greater depth, evaluating six ABSA tasks across seven languages. In contrast, prior studies typically focus on fewer languages and only a single compound ABSA task. We also offer practical guidance for selecting the most suitable approach to monolingual and cross-lingual ABSA, depending on task-specific constraints and real-world requirements. These recommendations enhances the flexibility of our study and provides actionable insights for practitioners seeking effective deployment strategies.\footnote{We publish our codebase and data at \url{https://github.com/biba10/Generative-Cross-lingual-ABSA}.} 

Compared to our earlier work~\cite{smid-etal-icaart-25}, this paper introduces three additional ABSA tasks (ATE, ACD, ACSA), one more language (Czech), broader combinations of source and target languages, an evaluation of different LLMs, and the introduction of multi-tasking models.

The rest of the paper is organized as follows. Section~\ref{sec:related} reviews the related work in cross-lingual and monolingual ABSA. Section~\ref{sec:methodology} describes our proposed methodology. Section~\ref{sec:experiments} details our experimental setup and the datasets utilized. Section~\ref{sec:results} presents the results and findings derived from our experiments. Section~\ref{sec:discussion} thoroughly discusses the methodology's implications and offers practical recommendations based on our results. Finally, Section~\ref{sec:conclusion} summarizes our conclusions and highlights the contributions of this study to the field of ABSA.

\section{Related Work}
\label{sec:related}

Modern monolingual ABSA is increasingly framed as a text generation task. Annotation-style and extraction-style paradigms show the viability of generative approaches~\cite{zhang2021generative}. Several works tackle sentiment quad prediction using natural language formats~\cite{zhang-etal-2021-aspect-sentiment}, multi-task frameworks with element prompts~\cite{gao-etal-2022-lego}, or tree-based tuple generation~\cite{mao-etal-2022-seq2path}. Others explore the impact of sentiment element order~\cite{hu-etal-2022-improving-aspect} or use multiple permutations to improve prediction~\cite{gou-etal-2023-mvp}.

Early cross-lingual ABSA methods focus on single tasks and typically rely on translation followed by label projection, either directly or using alignment tools like FastAlign~\cite{dyer-etal-2013-simple}. Data quality is improved through co-training~\cite{zhou2015clopinionminer}, instance selection~\cite{klinger-cimiano-2015-instance}, or constrained SMT~\cite{lambert-2015-aspect}, while cross-lingual embeddings enable language-agnostic learning~\cite{akhtar-etal-2018-solving,jebbara-cimiano-2019-zero,wang2018transition}. Recent work shifts towards E2E-ABSA using mPLMs like XLM-R~\cite{conneau-etal-2020-unsupervised} in combination with machine translation. Some approaches use only translated data~\cite{li2020unsupervised}, alignment-free label projection combined with distillation on unlabelled target language data~\cite{zhang-etal-2021-cross}, or contrastive learning~\cite{lin2023clxabsa}, but they face challenges such as translation noise, reliance on external tools, and limited language coverage. Adapting mPLMs effectively for cross-lingual ABSA remains an open problem.

LLMs like ChatGPT~\cite{chatgpt-2022} have shown strong performance in zero-shot settings, but their effectiveness in ABSA drops compared to smaller fine-tuned models~\cite{gou-etal-2023-mvp,zhang2023sentiment}. This gap grows with task complexity, especially in compound ABSA. Nevertheless, fine-tuned LLaMA-based models achieve state-of-the-art results on several compound tasks~\cite{smid-etal-2024-llama}. A key limitation of many LLMs is their English-centric pre-training.

\section{Methodology}
\label{sec:methodology}

This section describes our approach to handling the triplet task, i.e. the TASD task, which can be easily adapted for other tasks with slight modifications.

\subsection{Problem Statement}
Given a sentence as input, the objective is to predict all sentiment triplets $T={(a, c, p)}$, where each triplet comprises an aspect term ($a$), aspect category ($c$), and sentiment polarity ($p$). We adopt approaches from prior research~\cite{gou-etal-2023-mvp,hu-etal-2022-improving-aspect,zhang-etal-2021-aspect-sentiment} to transform sentiment elements ($a, c, p$) into natural language representations ($e_a, e_c, e_p$).

For the aspect term $a$, the representation $e_a$ is straightforward: it is the original aspect term, except in the case of a \textit{\quotes{NULL}} aspect term, which we replace with \textit{\quotes{it}}.

For the aspect category $c$, the original format is \texttt{\small ENTITY\#ATTRIBUTE}. We transform it into the representation $e_c$ by converting all letters to lowercase and replacing the \quotes{\#} with a space. For example, the aspect category \texttt{\small FOOD\#QUALITY} becomes \textit{\quotes{food quality}}.

For the sentiment polarity $p$, we use the following mapping function $\mathcal{P}_p(p)$ to obtain the representation $e_p$:
\begin{equation}
    \mathcal{P}_p(p) = 
    \begin{cases}
        \text{\textit{great}} & \text{if} \; p \: \text{is \textit{positive},}\\
        \text{\textit{ok}} & \text{if} \; p \: \text{is \textit{neutral},}\\
        \text{\textit{bad}} & \text{if} \; p \: \text{is \textit{negative}.}\\
    \end{cases}
\end{equation}

\subsection{Constructing Input and Output}
In crafting the inputs and outputs for our model, we use element markers~\cite{gou-etal-2023-mvp,hu-etal-2022-improving-aspect} to denote each sentiment element: {\sffamily[A]} for $e_a$, {\sffamily[C]} for $e_c$, and {\sffamily[P]} for $e_p$. These markers precede each element, collectively forming the target sequence. Additionally, we append these markers as a prompt to the input sequence $s$ to guide the model towards producing the correct output as $x = s \:|\: {\mathrm{\mathsf[A]}} \: {\mathrm{\mathsf[C]}} \: {\mathrm{\mathsf[P]}}$. Following the prioritization order $e_a > e_c > e_p$ recommended by a previous study~\cite{gou-etal-2023-mvp}, we create input-output pairs as shown below:\\
\hspace*{7pt}\textbf{\textit{Input ($x$)}:} Delicious tea but pricey soup | {\sffamily[A]} {\sffamily[C]} {\sffamily[P]} \\
\hspace*{7pt}\textbf{\textit{Output ($y$)}:} {\sffamily[A]} tea {\sffamily[C]} drinks quality {\sffamily[P]} great {\sffamily[;]} {\sffamily[A]} soup {\sffamily[C]} food prices {\sffamily[P]} bad

For sentences containing multiple sentiment tuples, we use the sequence {\sffamily[;]} to concatenate their target schemes into the final target sequence.

\subsection{Training}
\begin{figure}[ht!]
    \centering
    \includegraphics[width=\linewidth]{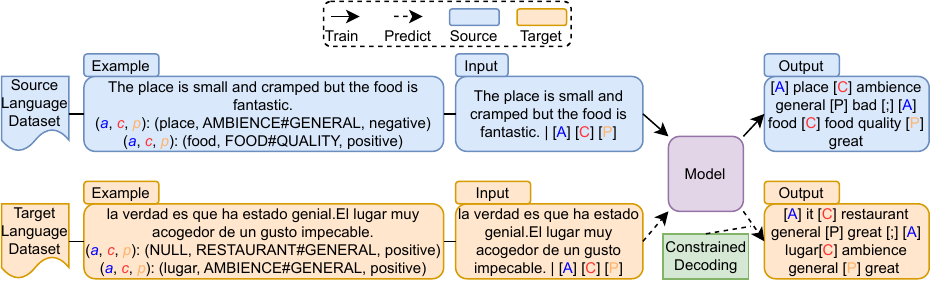}
    \caption{Overview of our method including input creation, training and generation process with expected output~\cite{smid-etal-icaart-25}.}
    \label{fig:training}
\end{figure}

We fine-tune a pre-trained sequence-to-sequence model using provided input-output pairs. Sequence-to-sequence models, also called encoder-decoder models, consist of two components: the encoder and the decoder. The encoder transforms input sequence $x$ into a contextualized encoded sequence $\vec{e}$. The decoder models the conditional probability distribution $P_{\vec{\Theta}}(y|\vec{e})$ of the target sequence $y$ based on the encoded input $\vec{e}$, where $\vec{\Theta}$ represents the model parameters. At each decoding step $i$, the decoder generates the output $y_i$ using previous outputs $y_0, \ldots, y_{i-1}$ and the encoded input $\vec{e}$. Given the input-target pair $(x, y)$ and model parameters $\vec{\Theta}$, initialized with the pre-trained weights, we further fine-tune the parameters to minimize the log-likelihood as
\begin{equation}
    \mathcal{L} = -\sum_{i=1}^n\log p_{\vec{\Theta}}(y_i|\vec{e},y_{<i}),
\end{equation}
where $n$ is the length of the target sequence $y$. Figure~\ref{fig:training} shows the example of input creation, training and generation process.

\subsubsection{Multi-Task Learning}
\begin{figure}[ht!]
    \centering
    \includegraphics[width=0.9\linewidth]{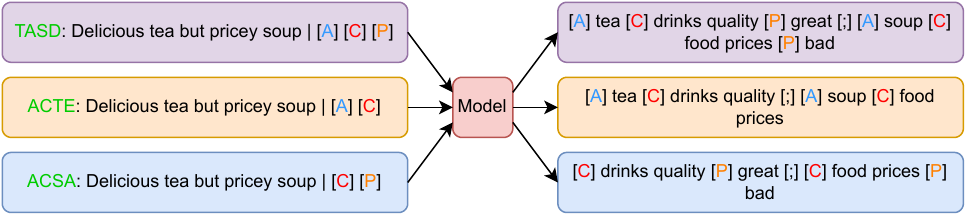}
    \caption{Example of multi-task learning. The model simultaneously addresses multiple ABSA tasks based on different input prompts.}
    \label{fig:multitask}
\end{figure}

We enable multi-task learning by modifying the prompt appended to the input sentence, allowing the model to predict multiple tasks simultaneously. Depending on the additional markers included in the input prompt, the model selects the task to perform. Figure~\ref{fig:multitask} illustrates an example of multi-task learning.

\subsection{Scheme-Guided Constrained Decoding}
In refining our model, we encountered issues where the fine-tuned model might not follow the desired output format, occasionally generate aspect terms in the source language rather than the target language or generate text not present in the original review. To address this, we introduce scheme-guided constrained decoding (CD)~\cite{constrained}, a method designed to ensure the generated elements align with their intended vocabulary sets by leveraging target schema information. This approach proves particularly beneficial in scenarios with limited training data in monolingual settings~\cite{gou-etal-2023-mvp}.

Constrained decoding solves common decoding problems, where the model searches the entire vocabulary for the subsequent token. By employing constrained decoding, we prevent the generation of undesired sequences that fail to meet our specifications. This technique operates dynamically, adjusting the list of potential candidate tokens based on the current state and checking the tokens individually to ensure a more controlled and precise generation process. For instance, when the current token is '{\sffamily[}', the subsequent token selection should be restricted to special terms: '{\sffamily A}', '{\sffamily C}', '{\sffamily P}', and '{\sffamily ;}'. Furthermore, constrained decoding monitors the generated output and current term, guiding the selection of subsequent tokens according to the criteria outlined in Table~\ref{tab:CD}.

\begin{table}[ht!]
    \centering
    \caption{Candidate lists of tokens for the TASD task. \texttt{<eos>} indicates the end of a sequence, and \quotes{$\ldots$} denotes arbitrary text.}
     \begin{adjustbox}{width=0.4\linewidth}
        \begin{tabular}{@{}ll@{}}
            \toprule
            \textbf{Generated output}                & \textbf{Candidate tokens} \\ \midrule
                                                    & [                         \\
            $\ldots$ {\sffamily [A} / {\sffamily [C} / {\sffamily [P} / {\sffamily [;} & {\sffamily ]}                         \\
            $\ldots$ {\sffamily [A]}                                   & Input sentence            \\
            $\ldots$ {\sffamily [C]}                                   & All categories            \\
            $\ldots$ {\sffamily [P]}                                   & great, ok, bad            \\
            $\ldots$ {\sffamily [A]} $\ldots$                                 & Input sentence, {\sffamily [}         \\
            $\ldots$ {\sffamily [C]} $\ldots$                                 & All categories, {\sffamily [}         \\
            $\ldots$ {\sffamily [P]} $\ldots$                                 & great, ok, bad, \texttt{<eos>} {\sffamily [}         \\
            $\ldots$ {\sffamily [A]} $\ldots$ {\sffamily [}                               & {\sffamily C}                         \\
            $\ldots$ {\sffamily [C]} $\ldots$ {\sffamily [}                               & {\sffamily P}                         \\
            $\ldots$ {\sffamily [P]} $\ldots$ {\sffamily [}                               & ;                         \\
            $\ldots$ {\sffamily [;]}                                   & {\sffamily [}                         \\
            $\ldots$ {\sffamily [;]} {\sffamily [}                                 & {\sffamily A}                         \\ \bottomrule
        \end{tabular}
    \end{adjustbox}
    \label{tab:CD}
\end{table}

For example, when the model generates an aspect term, constrained decoding limits the available tokens to only those present in the input (target) language sentence. Similarly, the available tokens are restricted to predefined aspect categories when generating an aspect category. The same principle applies to sentiment polarity, where the tokens can only come from permissible sentiment polarities. This method ensures that the generated output is consistent with the expected format and content.

Algorithm~\ref{algo:constrained} shows the pseudo-code of proposed CD algorithm.

\begin{algorithm}[ht!]
{\small
\KwData{Generated sequence, Input sentence tokens, Special token map}
\KwResult{Candidate tokens for the next step}
Get positions of \quotes{[} and \quotes{]} in the generated sequence\;

\uIf{no \quotes{[} tokens generated}{
    \Return \quotes{[}\;
}

Count \quotes{[} and \quotes{]} tokens and find last \quotes{[}\;

Get last generated token\;

\uIf{fewer \quotes{]} than \quotes{[} and last generated token is special}{
    \Return \quotes{]}\;
}

\uIf{last generated token is \quotes{[}}{
    \uIf{last special token is \quotes{;} or none}{
        \Return \quotes{A}\;
    }
    \uIf{last special token is \quotes{A}}{
        \Return \quotes{C}\;
    }
    \uIf{last special token is \quotes{C}}{
        \Return \quotes{P}\;
    }
    \uIf{last special token is \quotes{P}}{
        \Return \quotes{;}\;
    }
}

\uIf{last special token is \quotes{;}}{
    \Return \quotes{[}\;
}

Initialize result as an empty list\;

\uIf{last special token is \quotes{A}}{
    Add input sentence tokens and \quotes{it} to result\;
}

\uIf{last special token is \quotes{C}}{
    Add category tokens to result\;
}

\uIf{last special token is \quotes{P}}{
    Add sentiment tokens to result\;
}

\uIf{last generated token is not \quotes{]}}{
    Add \quotes{]} to result\;
    \uIf{last special token is \quotes{P}}{
       Add \quotes{$\langle$eos$\rangle$} to result\;
    }
}
\Return result\;
}
\caption{Proposed constrained decoding for the TASD task~\cite{smid-etal-icaart-25}.}
\label{algo:constrained}
\end{algorithm}

\subsection{Large Language Models Prompts}
\begin{figure}[ht!]
    \centering
    \includegraphics[width=\linewidth]{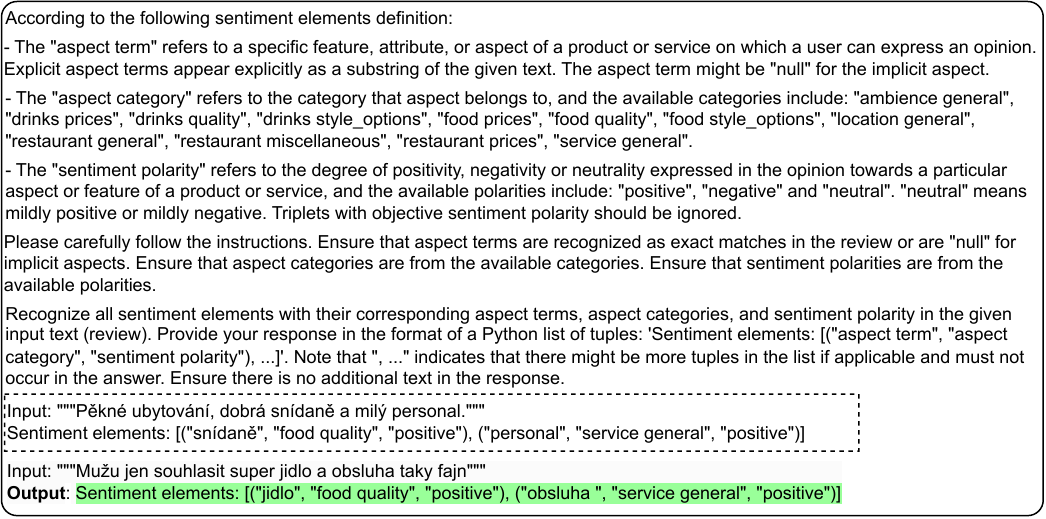}
    \caption{Prompt for the TASD task with example input, expected output in a green box, and three demonstrations in Czech enclosed in a dashed box. The demonstrations are used solely in few-shot scenarios~\cite{smid-etal-icaart-25}.}
    \label{fig:prompt_tasd}
\end{figure}

Figure~\ref{fig:prompt_tasd} illustrates the prompt for the TASD task with expected input, output, and few-shot demonstrations in Czech. The prompt is adaptable for various tasks by omitting the unnecessary sentiment element for the specific task, such as the sentiment polarity for the ACTE task. Few-shot examples are drawn from the first ten examples of the training dataset in the respective language for a fair evaluation. The distribution of labels in these examples is similar to the entire dataset, ensuring a random and representative sample.

\section{Experiment Setup}
\label{sec:experiments}
This section describes the datasets used in our experiments, the ABSA tasks we tackle, and the detailed experimental setup, including hyperparameters and evaluation metrics.

\subsection{Data \& Tasks}

For the experiments, we use the SemEval-2016 Task 5 dataset~\cite{pontiki-etal-2016-semeval}, which contains restaurant reviews in English (en), Spanish (es), French (fr), Dutch (nl), Russian (ru), and Turkish (tr). Each dataset is pre-split into training and test sets. We further divide the training data into a 9:1 ratio to obtain a validation set. Additionally, we use the CsRest-M dataset~\cite{smid-etal-2024-czech-dataset} that contains restaurant reviews in Czech (cs). This dataset is provided with training, validation and test splits. Table~\ref{tab:data_stats} shows the statistics for each dataset.

\begin{table*}[ht!]
    \centering
    \caption{Dataset statistics for each language. POS, NEG and NEU denote the number of positive, negative and neutral examples, respectively.}
    \begin{adjustbox}{width=\linewidth}
        \begin{tabular}{@{}llrrrrrrr@{}}
            \toprule
                                                    &              & \textbf{Cs}     & \textbf{En}  & \textbf{Es}   & \textbf{Fr}     & \textbf{Nl} & \textbf{Ru}   & \textbf{Tr} \\ \midrule
            \multirow{5}{*}{Train}                  & Sentences    & 2,151           & 1,800        & 1,863         & 1,559           & 1,549       & 3,289         & 1,108       \\
                                                    & Triplets     & 4,386           & 2,266        & 2,455         & 2,276           & 1,676       & 3,697         & 1,386       \\
                                                    & Categories   & 12              & 12           & 12            & 12              & 12          & 12            & 12          \\
                                                    & POS/NEG/NEU  & 2,663/1,338/385 & 1,503/672/91 & 1,736/607/112 & 1,045/1,092/139 & 969/584/124 & 2,805/641/250 & 746/521/119 \\
                                                    & NULL aspects & 961             & 569          & 700           & 694             & 513         & 821           & 135         \\\cdashlinelr{1-9}
            \multirow{5}{*}{Dev}  & Sentences    & 240             & 200          & 207           & 174             & 173         & 366           & 124         \\
                                                    & Triplets     & 483             & 241          & 265           & 254             & 184         & 392           & 149         \\
                                                    & Categories   & 12              & 11           & 11            & 12              & 11          & 12            & 10          \\
                                                    & POS/NEG/NEU  & 278/161/44      & 154/77/10    & 189/67/8      & 115/120/15      & 94/62/28    & 298/68/26     & 74/65/10    \\
                                                    & NULL aspects & 104             & 58           & 83            & 66              & 64          & 109           & 15          \\\cdashlinelr{1-9}
            \multirow{5}{*}{Test} & Sentences    & 798             & 676          & 881           & 694             & 575         & 1,209         & 144         \\
                                                    & Triplets     & 1,609           & 859          & 1,072         & 954             & 613         & 1,300         & 159         \\
                                                    & Categories   & 12              & 12           & 12            & 13              & 13          & 12            & 11          \\
                                                    & POS/NEG/NEU  & 972/497/140     & 611/204/44   & 750/274/48    & 441/434/79      & 369/211/33  & 870/321/103   & 104/49/6    \\
                                                    & NULL aspects & 342             & 209          & 341           & 236             & 219         & 325           & 0           \\ \bottomrule
        \end{tabular}
    \end{adjustbox}
    \label{tab:data_stats}
\end{table*}

We address six ABSA tasks: ATE, ACD, ACSA, E2E-ABSA, ACTE, and TASD.

\subsection{Experiment Details}
We employ two models, large mT5~\cite{xue-etal-2021-mt5} and large mBART~\cite{tang-etal-2021-multilingual}, from the Huggingface Transformers library\footnote{\url{https://github.com/huggingface/transformers}}~\cite{wolf-etal-2020-transformers}. The choice of the mT5 model follows prior English research~\cite{gao-etal-2022-lego,gou-etal-2023-mvp,hu-etal-2022-improving-aspect,zhang-etal-2021-aspect-sentiment,zhang2021generative} utilizing the monolingual T5 model~\cite{raffel2023exploring}. The inclusion of the mBART model aims to assess the robustness of our approach with a different backbone model. All experiments are conducted using an NVIDIA A40 with 48 GB GPU memory.

We maintain the same settings across all experiments, chosen based on consistent validation performance across all languages and tasks. We use a batch size 16 and conduct training for 20 epochs using a greedy search for decoding in all experiments. For the mT5 model, we set the learning rate to 1e-4 and utilize the Adafactor optimizer~\cite{shazeer2018adafactor}. For the mBART model, we set the learning rate to 1e-5 and employ the AdamW optimizer~\cite{loshchilov2019decoupled}. Given that all examples fit within the maximum length of 512, there is no need to trim the input to meet the maximum length requirements of the models. During the fine-tuning process, we update all model parameters. Multi-tasking models are fine-tuned on all six tasks simultaneously using the same hyperparameters as single-task models. Since mT5 outperformed mBART in preliminary experiments and to reduce the number of experiments, we use only mT5 for multi-tasking.

\subsection{Large Language Models}
We compare our method with several LLMs, -- LLaMA~2~\cite{touvron2023llama2}, LLaMA~3~\cite{dubey2024llama3herdmodels}, Orca~2~\cite{mitra2023orca}, and ChatGPT (\texttt{gpt-3.5-turbo})~\cite{chatgpt-2022}. Notably, we assess both the 13B and 7B versions of LLaMA~2 and Orca~2, as well as the 8B version of LLaMA~3. Our evaluation encompasses zero-shot and few-shot prompts across compound tasks. Moreover, we conduct instruction tuning for Orca~2, LLaMA~2, and LLaMA~3 in monolingual and cross-lingual settings.

For ChatGPT, we utilize the official paid API\footnote{\url{https://platform.openai.com/}}. For other models, which are open-source, unlike ChatGPT, we apply 4-bit quantization to fit the model into GPU memory. Preliminary results indicate that 4-bit quantization performs comparably to 8-bit quantization.

\subsubsection{Instruction Tuning}
We employ QLoRA~\cite{dettmers2023qlora} with 4-bit NormalFloat quantization to fine-tune the LLMs. This technique utilizes a quantized 4-bit frozen backbone LLM with a small set of learnable LoRA~\cite{hu2021lora} weights, enabling fine-tuning of LLMs on a single consumer GPU. Following the recommendations in the QLoRA paper, we adopt a batch size of 16, a constant learning rate of 2e-4, AdamW optimizer, and apply LoRA adapters on all linear transformer block layers with LoRA settings of $r=64$ and $\alpha=16$. Utilizing the zero-shot prompt (without demonstrations) shown in Figure~\ref{fig:prompt_tasd}, we fine-tune the model for up to 5 epochs, selecting the best-performing model based on validation loss.

\subsection{Evaluation Metrics}
The primary evaluation metric used is the micro F1 score, a standard metric in ABSA research~\cite{gou-etal-2023-mvp,zhang-etal-2021-aspect-sentiment,zhang-etal-2021-cross}. We define a predicted sentiment tuple as correct only if all its elements precisely match the gold tuple. Results are presented with a 95\% confidence interval derived from 5 runs with different random seeds.

\subsection{Compared Methods}
Where possible, we compare our method to existing cross-lingual works. This includes the ATE task and E2E-ABSA tasks. However, the compared methods for the ATE task involves older machine learning approaches, as mentioned in Section~\ref{sec:related}. For the E2E-ABSA tasks, the methods use encoder-based Transformer models. This lack of research on compound ABSA tasks is one of the motivations for this paper. 

\section{Results}
\label{sec:results}

This section presents results for six ABSA tasks using mT5 and mBART models, and four compound ABSA tasks using LLMs. In the tables, notation such as en$\rightarrow$cs indicates English as the source language and Czech as the target. We begin with simple ABSA tasks, followed by pair extraction tasks, and then provide a detailed analysis of the most complex task, TASD. Next, we present results across all LLMs, expanding on the earlier focus on ChatGPT and Orca~2~13B, which generally perform best. Finally, we discuss training and inference speeds for different models and provide an error analysis highlighting key challenges in the tasks.

\subsection{Simple ABSA Tasks}
\label{sub:res_simple}
Table~\ref{tab:res_simple_tasks} shows the results for simple ABSA tasks, highlighting the effectiveness of constrained decoding for the cross-lingual ATE task. This advantage arises from instances where the model correctly identifies the aspect term but in the source language instead of the target one. For example, the model might predict \textit{\quotes{place}} instead of the Czech \textit{\quotes{místo}} when Czech is the target language. Constrained decoding mitigates this problem effectively by restricting the available tokens for generation to only those from the input target language sentence. 
The results indicate that mT5 generally outperforms mBART and benefits more from constrained decoding. For the ACD task, such issues with generation do not occur since aspect categories are always predicted in English, resulting in no significant improvement from constrained decoding. Furthermore, constrained decoding does not significantly enhance monolingual results, as the model has no problems generating the aspect terms in a given source language.

\begin{table}[ht!]
\centering
\caption{Monolingual (for target languages) and cross-lingual F1 scores for for simple ABSA tasks. \textbf{Bold} indicates significant improvements with constrained decoding (CD) over without. The best cross-lingual result per task and language pair is \underline{underlined}. Asterisks (*) denote multi-tasking models. The compared works have the same data splits and task definition.}
\begin{adjustbox}{width=\linewidth}
\begin{tabular}{@{}llllcccccccccccc@{}}
\toprule
\textbf{Task}        & \textbf{Settings} & \textbf{Model}   &     & \textbf{En$\rightarrow$cs} & \textbf{En$\rightarrow$es} & \textbf{En$\rightarrow$fr} & \textbf{En$\rightarrow$nl} & \textbf{En$\rightarrow$ru} & \textbf{En$\rightarrow$tr} & \textbf{Cs$\rightarrow$en} & \textbf{Es$\rightarrow$en} & \textbf{Fr$\rightarrow$en} & \textbf{Nl$\rightarrow$en} & \textbf{Ru$\rightarrow$en} & \textbf{Tr$\rightarrow$en} \\ \midrule
\multirow{12}{*}{ACD} & \multirow{6}{*}{\rotatebox[origin=c]{90}{Monolingual}} & \multirow{4}{*}{mT5}   & Without CD  & 84.5$^{\pm0.4}$          & 82.3$^{\pm0.7}$          & 77.4$^{\pm0.4}$     & 81.8$^{\pm0.6}$          & 85.7$^{\pm0.4}$          & 80.7$^{\pm2.4}$          & 84.7$^{\pm0.6}$          & 84.7$^{\pm0.6}$          & 84.7$^{\pm0.6}$          & 84.7$^{\pm0.6}$          & 84.7$^{\pm0.6}$          & 84.7$^{\pm0.6}$     \\
                     & &                               & With CD & 84.2$^{\pm0.6}$          & 82.3$^{\pm0.6}$          & 76.9$^{\pm0.5}$     & 80.9$^{\pm0.6}$          & 85.8$^{\pm0.8}$          & 81.3$^{\pm0.7}$          & 84.8$^{\pm0.7}$          & 84.8$^{\pm0.7}$          & 84.8$^{\pm0.7}$          & 84.8$^{\pm0.7}$          & 84.8$^{\pm0.7}$          & 84.8$^{\pm0.7}$     \\
                     & &                               & Without CD* & 84.5$^{\pm0.4}$ & 82.3$^{\pm0.7}$ & 77.4$^{\pm0.4}$ & 81.8$^{\pm0.6}$ & 85.7$^{\pm0.4}$ & 80.7$^{\pm2.4}$ & 84.7$^{\pm0.6}$ & 84.7$^{\pm0.6}$ & 84.7$^{\pm0.6}$ & 84.7$^{\pm0.6}$ & 84.7$^{\pm0.6}$ & 84.7$^{\pm0.6}$\\
                     & &                               & With CD* & 84.2$^{\pm0.6}$ & 82.3$^{\pm0.6}$ & 76.9$^{\pm0.5}$ & 80.9$^{\pm0.6}$ & 85.8$^{\pm0.8}$ & 81.3$^{\pm0.7}$ & 84.8$^{\pm0.7}$ & 84.8$^{\pm0.7}$ & 84.8$^{\pm0.7}$ & 84.8$^{\pm0.7}$ & 84.8$^{\pm0.7}$ & 84.8$^{\pm0.7}$ \\
                    \cdashlinelr{3-16}
                      &  & \multirow{2}{*}{mBART} & Without CD   & 81.5$^{\pm0.4}$ & 79.6$^{\pm1.1}$ & 75.2$^{\pm0.6}$ & 77.9$^{\pm1.2}$ & 85.0$^{\pm1.3}$ & 74.9$^{\pm3.4}$ & 82.3$^{\pm1.2}$ & 82.3$^{\pm1.2}$ & 82.3$^{\pm1.2}$ & 82.3$^{\pm1.2}$ & 82.3$^{\pm1.2}$ & 82.3$^{\pm1.2}$\\
                      &  &  & With CD  & 81.6$^{\pm0.9}$ & 78.8$^{\pm1.1}$ & 74.0$^{\pm0.9}$ & 77.3$^{\pm0.8}$ & 83.9$^{\pm1.6}$ & 78.4$^{\pm2.0}$ & 83.2$^{\pm1.3}$ & 83.2$^{\pm1.3}$ & 83.2$^{\pm1.3}$ & 83.2$^{\pm1.3}$ & 83.2$^{\pm1.3}$ & 83.2$^{\pm1.3}$\\
                     \cdashlinelr{2-16}
                     & \multirow{6}{*}{\rotatebox[origin=c]{90}{Cross-lingual}} & \multirow{4}{*}{mT5}     & Without CD  & 77.2$^{\pm1.2}$          & 76.3$^{\pm0.7}$          & 73.9$^{\pm0.9}$     & 74.7$^{\pm0.3}$          & 80.2$^{\pm1.6}$          & 76.5$^{\pm1.6}$          & \underline{80.9}$^{\pm0.7}$          & \underline{80.5}$^{\pm0.2}$          & 81.5$^{\pm0.6}$          & 80.3$^{\pm0.6}$          & \underline{81.1}$^{\pm0.6}$          & \underline{79.3}$^{\pm1.0}$     \\
                     &   &                             & With CD & 77.3$^{\pm0.8}$          & 77.3$^{\pm0.3}$          & 74.8$^{\pm0.8}$     & 75.4$^{\pm1.4}$          & 80.2$^{\pm0.9}$          & 76.4$^{\pm1.6}$          & 80.8$^{\pm0.9}$          & 80.4$^{\pm0.8}$          & 81.3$^{\pm0.6}$          & \underline{80.7}$^{\pm0.5}$          & 80.9$^{\pm0.4}$          & 78.8$^{\pm0.6}$     \\
                      &  &                            & Without CD*    & 76.9$^{\pm1.1}$                      & 76.8$^{\pm0.6}$                      & \underline{75.9}$^{\pm1.3}$                      & 75.3$^{\pm1.2}$                      & 80.3$^{\pm0.6}$              & 76.0$^{\pm2.4}$              & 79.9$^{\pm0.8}$              & 80.1$^{\pm0.5}$              & 80.3$^{\pm1.1}$              & 80.5$^{\pm0.6}$              & 80.8$^{\pm0.5}$              & 78.2$^{\pm0.9}$              \\
                     &   &                           & With CD*   & \underline{77.8}$^{\pm0.9}$                      & \underline{77.7}$^{\pm1.0}$                      & 75.7$^{\pm0.5}$                      & \underline{75.6}$^{\pm1.2}$                      & \underline{80.8}$^{\pm0.7}$              & \underline{77.1}$^{\pm2.0}$              & 80.6$^{\pm0.6}$              & 80.1$^{\pm0.3}$              & \underline{81.6}$^{\pm0.8}$              & 80.4$^{\pm1.3}$              & 81.0$^{\pm0.7}$              & 78.5$^{\pm0.8}$              \\
                     \cdashlinelr{3-16}
                     & & \multirow{2}{*}{mBART}     & Without CD & 69.2$^{\pm2.2}$   & 69.7$^{\pm0.8}$   & 68.4$^{\pm0.9}$   & 68.9$^{\pm2.8}$   & 77.7$^{\pm1.6}$   & 60.6$^{\pm6.5}$ & 74.3$^{\pm2.4}$ & 75.7$^{\pm1.5}$ & 75.6$^{\pm2.5}$ & 77.0$^{\pm1.9}$ & 75.0$^{\pm1.4}$ & 72.9$^{\pm1.7}$ \\  
                     
                     & &     & With CD& 67.5$^{\pm1.7}$          & 70.7$^{\pm1.4}$          & 69.1$^{\pm1.1}$          & 69.5$^{\pm2.8}$          & 77.8$^{\pm1.5}$          & 63.5$^{\pm3.0}$ 
                     & 75.7$^{\pm2.0}$          & 76.9$^{\pm0.9}$          & 76.9$^{\pm1.2}$          & 77.4$^{\pm1.3}$          & 78.0$^{\pm1.7}$          & 74.7$^{\pm1.5}$\\
                     \midrule
\multirow{15}{*}{ATE} & \multirow{7}{*}{\rotatebox[origin=c]{90}{Monolingual}} & \multirow{4}{*}{mT5}   & Without CD  & 81.4$^{\pm0.4}$          & 80.4$^{\pm0.5}$          & 76.2$^{\pm0.8}$     & 80.0$^{\pm0.4}$          & 79.9$^{\pm0.6}$          & 68.1$^{\pm2.4}$          & 83.7$^{\pm0.3}$          & 83.7$^{\pm0.3}$          & 83.7$^{\pm0.3}$          & 83.7$^{\pm0.3}$          & 83.7$^{\pm0.3}$          & 83.7$^{\pm0.3}$     \\
                     &        &                        & With CD & 81.5$^{\pm0.7}$          & 80.7$^{\pm0.9}$          & 77.3$^{\pm0.4}$     & 72.8$^{\pm0.9}$          & 79.6$^{\pm0.6}$          & 68.0$^{\pm2.5}$          & 83.6$^{\pm0.8}$          & 83.6$^{\pm0.8}$          & 83.6$^{\pm0.8}$          & 83.6$^{\pm0.8}$          & 83.6$^{\pm0.8}$          & 83.6$^{\pm0.8}$     \\
                     &        &                        & Without CD* & 81.9$^{\pm0.7}$ & 81.2$^{\pm0.8}$ & 78.0$^{\pm1.1}$ & 81.9$^{\pm1.7}$ & 80.6$^{\pm0.2}$ & 69.6$^{\pm1.6}$ & 84.2$^{\pm0.7}$ & 84.2$^{\pm0.7}$ & 84.2$^{\pm0.7}$ & 84.2$^{\pm0.7}$ & 84.2$^{\pm0.7}$ & 84.2$^{\pm0.7}$\\
                     &        &                        & With CD* & 82.4$^{\pm0.7}$            & 80.9$^{\pm0.6}$ & 78.7$^{\pm1.2}$ & 75.1$^{\pm1.3}$ & 81.1$^{\pm0.5}$ & 68.6$^{\pm1.4}$  & 84.3$^{\pm0.2}$  & 84.3$^{\pm0.2}$  & 84.3$^{\pm0.2}$  & 84.3$^{\pm0.2}$  & 84.3$^{\pm0.2}$  & 84.3$^{\pm0.2}$             \\
                     \cdashlinelr{3-16}
                      &  & \multirow{2}{*}{mBART}   & Without CD & 79.2$^{\pm0.9}$ & 79.8$^{\pm1.2}$ & 74.6$^{\pm1.0}$ & 79.0$^{\pm1.3}$ & 78.6$^{\pm0.8}$ & 65.2$^{\pm5.0}$ & 80.9$^{\pm1.5}$ & 80.9$^{\pm1.5}$ & 80.9$^{\pm1.5}$ & 80.9$^{\pm1.5}$ & 80.9$^{\pm1.5}$ & 80.9$^{\pm1.5}$\\
                      &  &   & With CD& 77.7$^{\pm0.7}$ & 77.8$^{\pm2.1}$ & 72.0$^{\pm0.6}$ & 71.1$^{\pm1.7}$ & 75.4$^{\pm1.0}$ & 64.0$^{\pm1.4}$ & 80.8$^{\pm0.8}$ & 80.8$^{\pm0.8}$ & 80.8$^{\pm0.8}$ & 80.8$^{\pm0.8}$ & 80.8$^{\pm0.8}$ & 80.8$^{\pm0.8}$\\
                     \cdashlinelr{3-16}
                     &       &                          {\cite{jebbara-cimiano-2019-zero}} &      & --                            & 68.0\phantom{$^{\pm0.0}$}    & --                            & 60.0\phantom{$^{\pm0.0}$}    & 56.0\phantom{$^{\pm0.0}$}    & 48.0\phantom{$^{\pm0.0}$}    & 66.0\phantom{$^{\pm0.0}$}    & 66.0\phantom{$^{\pm0.0}$}    & 66.0\phantom{$^{\pm0.0}$}    & 66.0\phantom{$^{\pm0.0}$}    & 66.0\phantom{$^{\pm0.0}$}    & 66.0\phantom{$^{\pm0.0}$}    \\
\cdashlinelr{2-16}
                     & \multirow{8}{*}{\rotatebox[origin=c]{90}{Cross-lingual}}  & \multirow{4}{*}{mT5}   & Without CD  & 65.7$^{\pm0.2}$          & 63.4$^{\pm0.6}$          & 67.2$^{\pm0.5}$     & 62.3$^{\pm2.2}$          & 65.7$^{\pm1.3}$          & 52.2$^{\pm1.6}$          & 75.3$^{\pm1.0}$          & 73.1$^{\pm1.8}$          & 75.2$^{\pm1.0}$          & 70.8$^{\pm0.5}$          & 75.9$^{\pm0.1}$          & 65.9$^{\pm1.1}$     \\
                     &        &                        & With CD & \textbf{68.6}$^{\pm1.1}$ & \textbf{74.9}$^{\pm1.6}$ & 66.5$^{\pm0.6}$     & \textbf{65.2}$^{\pm0.6}$ & \textbf{70.1}$^{\pm1.3}$ & \textbf{\underline{54.5}}$^{\pm0.4}$ & \textbf{77.3}$^{\pm0.8}$ & \textbf{76.3}$^{\pm1.0}$ & \textbf{\underline{76.8}}$^{\pm0.3}$ & \textbf{\underline{77.5}}$^{\pm0.6}$ & \textbf{77.8}$^{\pm0.6}$ & 66.7$^{\pm0.6}$     \\
                     &       &                       & Without CD*    & 61.8$^{\pm1.5}$                      & 63.6$^{\pm1.0}$                      & 63.2$^{\pm1.2}$                      & 59.6$^{\pm1.0}$                      & 46.0$^{\pm2.3}$              & 41.2$^{\pm1.9}$              & 77.6$^{\pm0.4}$              & 76.6$^{\pm1.2}$              & 75.6$^{\pm1.2}$              & 71.0$^{\pm0.9}$              & 77.0$^{\pm1.0}$              & 69.2$^{\pm1.5}$              \\
                     &       &                       & With CD*   & \textbf{\underline{71.9}}$^{\pm1.4}$ & \textbf{\underline{78.4}}$^{\pm1.0}$ & \textbf{\underline{71.5}}$^{\pm0.8}$ & \textbf{\underline{67.9}}$^{\pm1.1}$ & \textbf{\underline{71.2}}$^{\pm1.3}$     & \textbf{51.9}$^{\pm2.2}$     & \underline{78.3}$^{\pm1.5}$              & \underline{77.2}$^{\pm1.2}$              & 76.1$^{\pm1.0}$              & \textbf{77.1}$^{\pm1.3}$     & \underline{78.8}$^{\pm0.8}$              & \underline{69.4}$^{\pm0.8}$              \\ 
                     \cdashlinelr{3-16}
                     &  & \multirow{2}{*}{mBART}   & Without CD & 62.1$^{\pm1.9}$   & 70.3$^{\pm2.0}$   & 60.9$^{\pm4.3}$   & 60.4$^{\pm2.8}$   & 66.5$^{\pm2.4}$   & 35.6$^{\pm4.2}$ & 76.2$^{\pm2.3}$ & 72.1$^{\pm1.2}$ & 73.4$^{\pm1.5}$ & 68.2$^{\pm2.3}$ & 77.1$^{\pm0.6}$ & 65.2$^{\pm2.3}$ \\
                     &  &   & With CD & 60.6$^{\pm1.9}$          & 71.5$^{\pm1.8}$          & 58.6$^{\pm0.6}$          & 59.7$^{\pm3.1}$          & 66.1$^{\pm2.5}$          & 38.6$^{\pm3.8}$
                      & 74.0$^{\pm1.4}$          & 73.4$^{\pm2.7}$          & 72.7$^{\pm1.1}$          & 71.7$^{\pm1.5}$          & 76.0$^{\pm1.4}$          & 63.4$^{\pm1.6}$ \\
                     \cdashlinelr{3-16}
                     &       &                          {\cite{wang2018transition}}  &    & --                            & 50.5\phantom{$^{\pm0.0}$}    & 50.0\phantom{$^{\pm0.0}$}    & --                            & --                            & --                            & --                            & 44.1\phantom{$^{\pm0.0}$}    & 50.3\phantom{$^{\pm0.0}$}    & --                            & --                            & --       \\                     
                     &         &                        {\cite{jebbara-cimiano-2019-zero}}  &      & --                            & 50.0\phantom{$^{\pm0.0}$}    & --                            & 46.0\phantom{$^{\pm0.0}$}    & 37.0\phantom{$^{\pm0.0}$}    & 17.0\phantom{$^{\pm0.0}$}    & --                            & 43.0\phantom{$^{\pm0.0}$}    & --                            & 45.0\phantom{$^{\pm0.0}$}    & 42.0\phantom{$^{\pm0.0}$}    & 33.0\phantom{$^{\pm0.0}$}    \\
                     
                     \bottomrule
\end{tabular}
\end{adjustbox}
\label{tab:res_simple_tasks}
\end{table}

The performance of multi-tasking models is noteworthy, as they often achieve the best overall cross-lingual results across various language pairs. Constrained decoding is especially beneficial for multi-tasking models in the ATE task, often yielding 10\% or more improvements over models without constrained decoding when English is the source language. In most cases, multi-tasking models with constrained decoding outperform or perform on par with models specialized in a single task, allowing for using a single model for multiple tasks without sacrificing performance.

When comparing monolingual and cross-lingual performance, cross-lingual results are generally about 4–6\% worse than monolingual results for the ACD task. The results for all other source languages are very similar when English is the target language. However, there are significant performance drops in some cross-lingual scenarios compared to monolingual results for the ATE task. For instance, using Turkish as both the source and target language results in a performance decrease of more than 10\% compared to monolingual results. Nevertheless, for most other language pairs, the cross-lingual results are only around 5\% worse than monolingual results, demonstrating the effectiveness of constrained decoding.

Additionally, the table includes comparisons with existing results for the ATE task~\cite{jebbara-cimiano-2019-zero,wang2018transition}. Our approach, especially with constrained decoding, significantly outperforms previous benchmarks, with improvements exceeding 20\% and even 30\% in most cases.

In summary, the results emphasize the effectiveness of constrained decoding in improving model performance in cross-lingual settings for aspect term detection. Multi-tasking models with constrained decoding are robust and capable, often surpassing specialized models.

\subsection{Pair Extraction ABSA Tasks}
Table~\ref{tab:res_pair} presents the results for the three pair-extraction ABSA tasks: E2E-ABSA, ACSA, and ACTE. These tasks differ in complexity, but several trends are consistent across them.

\begin{table}[ht!]
\centering
\caption{Monolingual (for target languages) and cross-lingual F1 scores for pair-wise tasks. \textbf{Bold} indicates significant improvements with constrained decoding (CD) over without. The best cross-lingual result per task and language pair is \underline{underlined}. Asterisks (*) denote multi-tasking models. ZS and FS stand for zero-shot and few-shot (10 examples). Compared E2E-ABSA baselines differ in data size and task definition.}
\begin{adjustbox}{width=\linewidth}
\begin{tabular}{@{}lllllcccccccccccc@{}}
\toprule
              \textbf{Task}  &    \multicolumn{2}{c}{\textbf{Setup}}                      & \textbf{Model}       &          & \textbf{En$\rightarrow$cs} & \textbf{En$\rightarrow$es}         & \textbf{En$\rightarrow$fr} & \textbf{En$\rightarrow$nl}         & \textbf{En$\rightarrow$ru} & \textbf{En$\rightarrow$tr} & \textbf{Cs$\rightarrow$en} & \textbf{Es$\rightarrow$en} & \textbf{Fr$\rightarrow$en} & \textbf{Nl$\rightarrow$en} & \textbf{Ru$\rightarrow$en} & \textbf{Tr$\rightarrow$en} \\ \midrule
\multirow{21}{*}{\rotatebox[origin=c]{90}{E2E-ABSA}} & \multirow{11}{*}{\rotatebox[origin=c]{90}{Monolingual}} & \multirow{2}{*}{\rotatebox[origin=c]{90}{ZS}}     & \multicolumn{2}{l}{ChatGPT}     & 44.6\phantom{$^{\pm0.0}$}    & 42.4\phantom{$^{\pm0.0}$}            & 37.1\phantom{$^{\pm0.0}$}    & 31.7\phantom{$^{\pm0.0}$}            & 35.1\phantom{$^{\pm0.0}$}    & 37.9\phantom{$^{\pm0.0}$}    & 55.5\phantom{$^{\pm0.0}$}    & 55.5\phantom{$^{\pm0.0}$}    & 55.5\phantom{$^{\pm0.0}$}    & 55.5\phantom{$^{\pm0.0}$}    & 55.5\phantom{$^{\pm0.0}$}    & 55.5\phantom{$^{\pm0.0}$}    \\
                    & &                                & \multicolumn{2}{l}{Orca~2~13B}  & 22.7\phantom{$^{\pm0.0}$}    & 27.4\phantom{$^{\pm0.0}$}            & 22.5\phantom{$^{\pm0.0}$}    & 18.3\phantom{$^{\pm0.0}$}            & 20.0\phantom{$^{\pm0.0}$}    & 19.2\phantom{$^{\pm0.0}$}    & 42.2\phantom{$^{\pm0.0}$}    & 42.2\phantom{$^{\pm0.0}$}    & 42.2\phantom{$^{\pm0.0}$}    & 42.2\phantom{$^{\pm0.0}$}    & 42.2\phantom{$^{\pm0.0}$}    & 42.2\phantom{$^{\pm0.0}$} \\  \cdashlinelr{3-17} 
                    & & \multirow{2}{*}{\rotatebox[origin=c]{90}{FS}} & \multicolumn{2}{l}{ChatGPT}     & 54.8\phantom{$^{\pm0.0}$}    & 59.1\phantom{$^{\pm0.0}$}            & 51.7\phantom{$^{\pm0.0}$}    & 51.6\phantom{$^{\pm0.0}$}            & 51.9\phantom{$^{\pm0.0}$}    & 42.7\phantom{$^{\pm0.0}$}    & 62.2\phantom{$^{\pm0.0}$}    & 62.2\phantom{$^{\pm0.0}$}    & 62.2\phantom{$^{\pm0.0}$}    & 62.2\phantom{$^{\pm0.0}$}    & 62.2\phantom{$^{\pm0.0}$}    & 62.2\phantom{$^{\pm0.0}$}    \\
                    & &                                & \multicolumn{2}{l}{Orca~2~13B}  & 41.0\phantom{$^{\pm0.0}$}    & 50.6\phantom{$^{\pm0.0}$}            & 43.9\phantom{$^{\pm0.0}$}    & 46.1\phantom{$^{\pm0.0}$}            & 38.7\phantom{$^{\pm0.0}$}    & 32.0\phantom{$^{\pm0.0}$}    & 62.4\phantom{$^{\pm0.0}$}    & 62.4\phantom{$^{\pm0.0}$}    & 62.4\phantom{$^{\pm0.0}$}    & 62.4\phantom{$^{\pm0.0}$}    & 62.4\phantom{$^{\pm0.0}$}    & 62.4\phantom{$^{\pm0.0}$}    \\ \cdashlinelr{3-17}
                    & & \multirow{7}{*}{\rotatebox[origin=c]{90}{Fine-tuning}}   
                                                   & \multicolumn{2}{l}{Orca~2~13B}  & 71.8$^{\pm1.1}$              & 74.8$^{\pm1.0}$                      & 69.8$^{\pm2.3}$              & 76.2$^{\pm0.3}$                      & 72.2$^{\pm1.2}$              & 52.2$^{\pm1.6}$              & 82.3$^{\pm0.6}$              & 82.3$^{\pm0.6}$              & 82.3$^{\pm0.6}$              & 82.3$^{\pm0.6}$              & 82.3$^{\pm0.6}$              & 82.3$^{\pm0.6}$              \\
                  \cdashlinelr{4-17}
                    & &                                & \multirow{4}{*}{mT5} & Without CD   & 73.4$^{\pm0.8}$              & 74.4$^{\pm0.6}$                      & 69.9$^{\pm0.5}$              & 71.6$^{\pm1.0}$                      & 72.4$^{\pm0.2}$              & 60.1$^{\pm1.7}$              & 77.7$^{\pm0.4}$              & 77.7$^{\pm0.4}$              & 77.7$^{\pm0.4}$              & 77.7$^{\pm0.4}$              & 77.7$^{\pm0.4}$              & 77.7$^{\pm0.4}$              \\
                    & &                                &                      & With CD  & 73.5$^{\pm0.4}$              & 75.3$^{\pm0.6}$                      & 69.8$^{\pm1.4}$              & 67.0$^{\pm0.4}$                      & 72.2$^{\pm0.4}$              & 60.7$^{\pm1.1}$              & 77.4$^{\pm0.5}$              & 77.4$^{\pm0.5}$              & 77.4$^{\pm0.5}$              & 77.4$^{\pm0.5}$              & 77.4$^{\pm0.5}$              & 77.4$^{\pm0.5}$              \\
                    & &                                &                      & Without CD* &  73.8$^{\pm0.5}$ & 75.3$^{\pm0.6}$ & 69.4$^{\pm0.8}$ & 73.2$^{\pm1.4}$ & 72.2$^{\pm0.4}$ & 63.4$^{\pm1.0}$ & 78.1$^{\pm1.1}$ & 78.1$^{\pm1.1}$ & 78.1$^{\pm1.1}$ & 78.1$^{\pm1.1}$ & 78.1$^{\pm1.1}$ & 78.1$^{\pm1.1}$\\
                    & &                                &                      & With CD*  & 74.6$^{\pm0.5}$ & 74.9$^{\pm1.3}$ & 69.9$^{\pm0.9}$ & 67.4$^{\pm1.3}$             & 73.2$^{\pm0.7}$             & 63.4$^{\pm1.7}$ & 78.0$^{\pm0.8}$ & 78.0$^{\pm0.8}$ & 78.0$^{\pm0.8}$ & 78.0$^{\pm0.8}$ & 78.0$^{\pm0.8}$ & 78.0$^{\pm0.8}$ \\
                    \cdashlinelr{4-17}
                    & &                                & \multirow{2}{*}{mBART} & Without CD   & 69.1$^{\pm0.3}$ & 73.0$^{\pm0.5}$ & 66.4$^{\pm1.1}$ & 68.9$^{\pm1.2}$ & 68.7$^{\pm1.6}$ & 56.0$^{\pm2.7}$ & 74.3$^{\pm1.4}$ & 74.3$^{\pm1.4}$ & 74.3$^{\pm1.4}$ & 74.3$^{\pm1.4}$ & 74.3$^{\pm1.4}$ & 74.3$^{\pm1.4}$\\
                    & &                                &  & Without CD   & 68.8$^{\pm0.8}$ & 71.9$^{\pm1.3}$ & 64.0$^{\pm1.7}$ & 61.6$^{\pm1.0}$ & 66.2$^{\pm1.1}$ & 54.4$^{\pm2.3}$ & 72.0$^{\pm1.4}$ & 72.0$^{\pm1.4}$ & 72.0$^{\pm1.4}$ & 72.0$^{\pm1.4}$ & 72.0$^{\pm1.4}$ & 72.0$^{\pm1.4}$\\
                    \cmidrule{2-17}
& \multirow{10}{*}{\rotatebox[origin=c]{90}{Cross-lingual}}  &  \multirow{10}{*}{\rotatebox[origin=c]{90}{Fine-tuning}}           
                                                    & \multicolumn{2}{l}{Orca~2~13B}  & 57.6$^{\pm0.4}$              & 69.5$^{\pm0.6}$                      & \underline{65.6}$^{\pm0.7}$              & 61.2$^{\pm0.3}$                      & 58.9$^{\pm1.7}$              & 33.8$^{\pm0.8}$              & \underline{76.5}$^{\pm0.6}$              & \underline{72.0}$^{\pm0.2}$              & \underline{75.6}$^{\pm0.3}$              & 66.3$^{\pm0.2}$              & \underline{79.0}$^{\pm0.6}$              & \underline{67.5}$^{\pm1.1}$              \\
                    \cdashlinelr{4-17}
                    & &                                & \multirow{4}{*}{mT5} & Without CD   & 57.3$^{\pm1.4}$              & 59.2$^{\pm0.5}$                      & 57.8$^{\pm1.2}$              & 57.1$^{\pm0.9}$                      & 56.4$^{\pm2.1}$              & 44.4$^{\pm1.4}$              & 68.8$^{\pm1.2}$              & 65.4$^{\pm1.0}$              & 69.1$^{\pm0.9}$              & 63.3$^{\pm0.5}$              & 68.0$^{\pm1.7}$              & 59.4$^{\pm1.1}$              \\
                    & &                                &                      & With CD  & \textbf{62.4}$^{\pm1.6}$     & \textbf{69.3}$^{\pm1.0}$             & \textbf{61.1}$^{\pm1.2}$     & \textbf{60.8}$^{\pm0.3}$             & \textbf{\underline{63.7}}$^{\pm1.3}$     & \textbf{\underline{48.9}}$^{\pm1.4}$     & 68.7$^{\pm0.9}$              & \textbf{68.9}$^{\pm1.1}$     & 68.3$^{\pm0.7}$              & \textbf{\underline{69.1}}$^{\pm0.5}$     & \textbf{70.6}$^{\pm0.7}$     & \textbf{61.5}$^{\pm0.7}$     \\
                    & &                                &                      & Without CD*  & 54.7$^{\pm1.3}$              & 58.2$^{\pm0.9}$                      & 55.1$^{\pm1.3}$              & 53.8$^{\pm1.9}$                      & 40.2$^{\pm1.9}$              & 38.2$^{\pm2.7}$              & 44.3$^{\pm2.9}$              & 44.7$^{\pm1.4}$              & 59.3$^{\pm1.7}$              & 54.2$^{\pm1.9}$              & 35.7$^{\pm1.0}$              & 42.9$^{\pm1.0}$              \\
                    & &                                &                      & With CD* & \textbf{\underline{64.0}}$^{\pm1.0}$     & \textbf{\underline{72.0}}$^{\pm1.0}$ & \textbf{62.7}$^{\pm1.4}$     & \textbf{62.2}$^{\pm1.0}$ & \textbf{63.4}$^{\pm0.9}$     & \textbf{47.0}$^{\pm0.6}$     & \textbf{65.0}$^{\pm1.7}$     & \textbf{64.9}$^{\pm1.3}$     & \textbf{67.3}$^{\pm0.8}$     & \textbf{67.9}$^{\pm1.6}$     & \textbf{58.9}$^{\pm2.7}$     & \textbf{58.5}$^{\pm1.7}$     \\
                    \cdashlinelr{4-17}
                    & &                                & \multirow{2}{*}{mBART} & Without CD

                    & 51.5$^{\pm3.3}$   & 61.1$^{\pm2.6}$   & 49.4$^{\pm3.8}$   & 51.6$^{\pm2.7}$   & 57.1$^{\pm1.4}$   & 31.6$^{\pm3.9}$  & 64.2$^{\pm2.2}$ & 64.5$^{\pm1.6}$ & 64.9$^{\pm3.2}$ & 59.5$^{\pm1.8}$ & 68.4$^{\pm1.2}$ & 57.6$^{\pm0.8}$ \\
                    &  &                                & & With CD                    
                    & 48.8$^{\pm2.7}$          & 61.7$^{\pm2.7}$          & 49.2$^{\pm4.1}$          & 50.1$^{\pm3.5}$          & 57.8$^{\pm1.8}$          & 30.3$^{\pm3.0}$ 
                    & 65.0$^{\pm1.3}$          & 64.9$^{\pm2.4}$          & 63.4$^{\pm1.1}$          & \textbf{64.1}$^{\pm1.3}$ & 68.5$^{\pm0.7}$          & 56.7$^{\pm0.8}$ \\
                    
                    \cdashlinelr{4-17}
                    & &                                & \multicolumn{2}{l}{\cite{zhang-etal-2021-cross}}            & --                            & 69.2\phantom{$^{\pm0.0}$}                                 & 61.0\phantom{$^{\pm0.0}$}                         & \underline{63.7}\phantom{$^{\pm0.0}$}                                 & 62.0\phantom{$^{\pm0.0}$}                         & --                            & --                            & --                            & --                            & --                            & --                            & --                           

                   \\
                    & &                                & \multicolumn{2}{l}{\cite{li2020unsupervised}}            & --                            & 67.1\phantom{$^{\pm0.0}$}                                 & 56.4\phantom{$^{\pm0.0}$}                         & 59.0\phantom{$^{\pm0.0}$}                                 & 56.8\phantom{$^{\pm0.0}$}                         & 46.2\phantom{$^{\pm0.0}$}                            & --                            & --                            & --                            & --                            & --                            & --                            \\
                    & &                                & \multicolumn{2}{l}{\cite{lin2023clxabsa}}            & --                            & 61.6\phantom{$^{\pm0.0}$}                                 & 49.5\phantom{$^{\pm0.0}$}                         & 51.0\phantom{$^{\pm0.0}$}                                 & 50.8\phantom{$^{\pm0.0}$}                         & --                            & --                            & --                            & --                            & --                            & --                            & --                            \\                              
                     \midrule
\multirow{22}{*}{\rotatebox[origin=c]{90}{ACSA}} & \multirow{11}{*}{\rotatebox[origin=c]{90}{Monolingual}}   & \multirow{2}{*}{\rotatebox[origin=c]{90}{ZS}}   & \multicolumn{2}{l}{ChatGPT}     & 57.3\phantom{$^{\pm0.0}$}  & 60.9\phantom{$^{\pm0.0}$}  & 58.1\phantom{$^{\pm0.0}$}  & 56.0\phantom{$^{\pm0.0}$}  & 55.8\phantom{$^{\pm0.0}$}  & 54.6\phantom{$^{\pm0.0}$}  & 60.3\phantom{$^{\pm0.0}$}   & 60.3\phantom{$^{\pm0.0}$}   & 60.3\phantom{$^{\pm0.0}$}   & 60.3\phantom{$^{\pm0.0}$}   & 60.3\phantom{$^{\pm0.0}$}  & 60.3\phantom{$^{\pm0.0}$}   \\
                                                       &  &                                                        & \multicolumn{2}{l}{Orca~2~13B}  & 49.5\phantom{$^{\pm0.0}$}  & 50.4\phantom{$^{\pm0.0}$}  & 46.3\phantom{$^{\pm0.0}$}  & 47.6\phantom{$^{\pm0.0}$}  & 51.0\phantom{$^{\pm0.0}$}  & 33.0\phantom{$^{\pm0.0}$}  & 59.6\phantom{$^{\pm0.0}$}   & 59.6\phantom{$^{\pm0.0}$}   & 59.6\phantom{$^{\pm0.0}$}   & 59.6\phantom{$^{\pm0.0}$}   & 59.6\phantom{$^{\pm0.0}$}  & 59.6\phantom{$^{\pm0.0}$}   \\
                                                         \cdashlinelr{3-17}
                                     &                      & \multirow{2}{*}{\rotatebox[origin=c]{90}{FS}}    & \multicolumn{2}{l}{ChatGPT}     & 61.6\phantom{$^{\pm0.0}$}  & 67.6\phantom{$^{\pm0.0}$}  & 62.2\phantom{$^{\pm0.0}$}  & 63.9\phantom{$^{\pm0.0}$}  & 64.9\phantom{$^{\pm0.0}$}  & 58.8\phantom{$^{\pm0.0}$}  & 64.7\phantom{$^{\pm0.0}$}   & 64.7\phantom{$^{\pm0.0}$}   & 64.7\phantom{$^{\pm0.0}$}   & 64.7\phantom{$^{\pm0.0}$}   & 64.7\phantom{$^{\pm0.0}$}  & 64.7\phantom{$^{\pm0.0}$}   \\
                                      &                     &                                                        & \multicolumn{2}{l}{Orca~2~13B}  & 63.4\phantom{$^{\pm0.0}$}  & 67.0\phantom{$^{\pm0.0}$}  & 55.7\phantom{$^{\pm0.0}$}  & 61.1\phantom{$^{\pm0.0}$}  & 59.8\phantom{$^{\pm0.0}$}  & 45.6\phantom{$^{\pm0.0}$}  & 65.2\phantom{$^{\pm0.0}$}   & 65.2\phantom{$^{\pm0.0}$}   & 65.2\phantom{$^{\pm0.0}$}   & 65.2\phantom{$^{\pm0.0}$}   & 65.2\phantom{$^{\pm0.0}$}  & 65.2\phantom{$^{\pm0.0}$}   \\
                                                         \cdashlinelr{3-17}
                                       &                    & \multirow{7}{*}{\rotatebox[origin=c]{90}{Fine-tuning}} 
                                                         & \multicolumn{2}{l}{Orca~2~13B}  & 75.4$^{\pm0.6}$            & 80.4$^{\pm0.4}$            & 78.1$^{\pm0.4}$            & 76.1$^{\pm0.7}$            & 80.2$^{\pm0.6}$            & 69.3$^{\pm0.8}$            & 84.3$^{\pm0.9}$             & 84.3$^{\pm0.9}$             & 84.3$^{\pm0.9}$             & 84.3$^{\pm0.9}$             & 84.3$^{\pm0.9}$            & 84.3$^{\pm0.9}$             \\
                                                         \cdashlinelr{4-17}
                                       &                    &                                                        & \multirow{4}{*}{mT5} & Without CD   & 76.6$^{\pm0.2}$            & 77.1$^{\pm0.2}$            & 69.2$^{\pm0.8}$            & 74.1$^{\pm0.3}$            & 78.0$^{\pm0.8}$            & 74.2$^{\pm1.5}$            & 78.6$^{\pm1.3}$             & 78.6$^{\pm1.3}$             & 78.6$^{\pm1.3}$             & 78.6$^{\pm1.3}$             & 78.6$^{\pm1.3}$            & 78.6$^{\pm1.3}$             \\
                                        &                   &                                                        &                      & With CD  & 76.5$^{\pm1.1}$            & 77.4$^{\pm0.6}$            & 69.0$^{\pm0.7}$            & 74.3$^{\pm0.9}$            & 77.7$^{\pm0.5}$            & 74.4$^{\pm4.6}$            & 78.0$^{\pm1.4}$             & 78.0$^{\pm1.4}$             & 78.0$^{\pm1.4}$             & 78.0$^{\pm1.4}$             & 78.0$^{\pm1.4}$            & 78.0$^{\pm1.4}$             \\                         &                               &                           &                      &
                                                         Without CD* & 75.9$^{\pm0.5}$ & 76.8$^{\pm0.5}$ & 69.5$^{\pm0.7}$ & 74.1$^{\pm0.7}$ & 77.4$^{\pm0.5}$ & 74.1$^{\pm1.8}$ & 79.0$^{\pm0.9}$ & 79.0$^{\pm0.9}$ & 79.0$^{\pm0.9}$ & 79.0$^{\pm0.9}$ & 79.0$^{\pm0.9}$ & 79.0$^{\pm0.9}$ \\                         &  &                                                        &                      & With CD* & 76.5$^{\pm0.7}$ & 76.7$^{\pm0.8}$ & 68.9$^{\pm0.6}$ & 75.2$^{\pm1.5}$             & 77.6$^{\pm0.5}$             & 75.0$^{\pm1.8}$ & 78.3$^{\pm1.0}$ & 78.3$^{\pm1.0}$ & 78.3$^{\pm1.0}$ & 78.3$^{\pm1.0}$ & 78.3$^{\pm1.0}$ & 78.3$^{\pm1.0}$\\ \cdashlinelr{4-17}
                                          &                 &                                                        & \multirow{2}{*}{mBART} & Without CD & 72.6$^{\pm1.3}$ & 73.2$^{\pm1.4}$ & 65.0$^{\pm0.9}$ & 70.2$^{\pm1.9}$ & 73.3$^{\pm1.2}$ & 66.1$^{\pm4.0}$ & 74.7$^{\pm1.7}$ & 74.7$^{\pm1.7}$ & 74.7$^{\pm1.7}$ & 74.7$^{\pm1.7}$ & 74.7$^{\pm1.7}$ & 74.7$^{\pm1.7}$ \\
                                          &                 &                                                        &  & With CD& 71.5$^{\pm0.8}$  & 73.3$^{\pm0.8}$ & 63.7$^{\pm1.0}$ & 68.3$^{\pm2.9}$ & 72.8$^{\pm1.0}$ & 64.4$^{\pm5.1}$ & 74.9$^{\pm1.7}$ & 74.9$^{\pm1.7}$ & 74.9$^{\pm1.7}$ & 74.9$^{\pm1.7}$ & 74.9$^{\pm1.7}$ & 74.9$^{\pm1.7}$\\
                                                         
                                                         \cmidrule{2-17}
 & \multirow{7}{*}{\rotatebox[origin=c]{90}{Cross-lingual}} &  \multirow{7}{*}{\rotatebox[origin=c]{90}{Fine-tuning}} 
                                                         & \multicolumn{2}{l}{Orca~2~13B}  & \underline{70.6}$^{\pm0.1}$            & \underline{74.8}$^{\pm0.7}$            & \underline{73.8}$^{\pm0.5}$            & 70.8$^{\pm0.4}$            & \underline{75.6}$^{\pm1.6}$            & 62.9$^{\pm2.1}$            & 79.6$^{\pm0.2}$             & \underline{81.4}$^{\pm0.3}$             & \underline{83.8}$^{\pm0.7}$             & \underline{81.0}$^{\pm0.8}$             & \underline{82.3}$^{\pm0.4}$            & \underline{80.5}$^{\pm0.2}$             \\
                                                         \cdashlinelr{4-17}
                                      &                     &                                                        & \multirow{4}{*}{mT5} & Without CD   & 68.0$^{\pm1.1}$            & 71.1$^{\pm0.8}$            & 63.7$^{\pm0.7}$            & 70.4$^{\pm0.7}$            & 71.5$^{\pm1.1}$            & 70.5$^{\pm2.7}$            & 73.8$^{\pm1.1}$             & 73.5$^{\pm0.2}$             & 74.2$^{\pm0.6}$             & 72.1$^{\pm0.7}$             & 74.0$^{\pm0.7}$            & 71.3$^{\pm1.2}$             \\
                                       &                    &                                                        &                      & With CD  & 67.7$^{\pm1.1}$            & 71.0$^{\pm1.1}$            & 64.4$^{\pm1.3}$            & 70.4$^{\pm1.1}$            & 71.4$^{\pm0.3}$            & \underline{70.7}$^{\pm3.0}$            & 72.9$^{\pm0.3}$             & 73.5$^{\pm0.5}$             & 74.7$^{\pm0.8}$             & 72.9$^{\pm1.1}$             & 74.2$^{\pm0.3}$            & 71.8$^{\pm0.8}$             \\
                                        &                   &                                                        &                      & Without CD*  & 67.4$^{\pm1.1}$            & 70.1$^{\pm0.9}$            & 64.8$^{\pm1.9}$            & 68.9$^{\pm0.6}$            & 71.1$^{\pm0.9}$            & 69.3$^{\pm1.9}$            & 76.0$^{\pm0.5}$ & 76.2$^{\pm1.6}$ & 76.6$^{\pm1.4}$ & 75.2$^{\pm1.3}$ & 75.9$^{\pm1.2}$            & 74.0$^{\pm1.4}$ \\
                                         &                  &                                                        &                      & With CD* & 68.2$^{\pm0.4}$            & 70.9$^{\pm1.6}$            & 65.2$^{\pm1.1}$            & 69.2$^{\pm0.7}$            & 71.3$^{\pm1.1}$            & 69.8$^{\pm2.7}$            & 75.2$^{\pm0.8}$ & 75.9$^{\pm1.0}$ & 76.7$^{\pm1.1}$ & 76.2$^{\pm0.9}$ & 74.5$^{\pm1.1}$            & 75.1$^{\pm0.4}$ \\ 
                                                         \cdashlinelr{4-17}
                                          &                &                                                        & \multirow{2}{*}{mBART} & Without CD
                                                         & 55.2$^{\pm1.4}$   & 61.7$^{\pm2.5}$   & 53.7$^{\pm2.4}$   & 58.4$^{\pm2.1}$   & 65.6$^{\pm1.9}$   & 50.7$^{\pm2.6}$ & 66.3$^{\pm2.3}$ & 67.1$^{\pm2.0}$ & 66.7$^{\pm1.6}$ & 66.7$^{\pm1.8}$ & 65.0$^{\pm1.3}$ & 63.8$^{\pm2.5}$ \\
                                        &                   &                                                        &  & With CD
                                                         
                                                         & 53.8$^{\pm4.0}$          & 63.0$^{\pm2.3}$          & 53.0$^{\pm1.8}$          & 59.4$^{\pm2.0}$          & 66.4$^{\pm1.7}$          & 49.3$^{\pm2.7}$ 
                                                           & 67.3$^{\pm2.1}$          & 68.3$^{\pm1.8}$          & \textbf{69.8}$^{\pm1.3}$ & \textbf{69.6}$^{\pm0.5}$ & \textbf{68.6}$^{\pm0.8}$ & 64.6$^{\pm1.4}$ \\
                                                         \midrule
\multirow{22}{*}{\rotatebox[origin=c]{90}{ACTE}} &\multirow{11}{*}{\rotatebox[origin=c]{90}{Monolingual}}  & \multirow{2}{*}{\rotatebox[origin=c]{90}{ZS}}   & \multicolumn{2}{l}{ChatGPT}     & 26.3\phantom{$^{\pm0.0}$}  & 30.1\phantom{$^{\pm0.0}$}  & 27.8\phantom{$^{\pm0.0}$}  & 20.9\phantom{$^{\pm0.0}$}  & 26.4\phantom{$^{\pm0.0}$}  & 26.3\phantom{$^{\pm0.0}$}  & 31.7\phantom{$^{\pm0.0}$}  & 31.7\phantom{$^{\pm0.0}$}  & 31.7\phantom{$^{\pm0.0}$}  & 31.7\phantom{$^{\pm0.0}$}  & 31.7\phantom{$^{\pm0.0}$}  & 31.7\phantom{$^{\pm0.0}$}  \\
                                    &                      &                                                        & \multicolumn{2}{l}{Orca~2~13B}  & 13.7\phantom{$^{\pm0.0}$}  & 15.3\phantom{$^{\pm0.0}$}  & 14.1\phantom{$^{\pm0.0}$}  & 10.1\phantom{$^{\pm0.0}$}  & 12.4\phantom{$^{\pm0.0}$}  & 10.4\phantom{$^{\pm0.0}$}  & 19.3\phantom{$^{\pm0.0}$}  & 19.3\phantom{$^{\pm0.0}$}  & 19.3\phantom{$^{\pm0.0}$}  & 19.3\phantom{$^{\pm0.0}$}  & 19.3\phantom{$^{\pm0.0}$}  & 19.3\phantom{$^{\pm0.0}$}  \\
                                                         \cdashlinelr{3-17}
                                   &                       & \multirow{2}{*}{\rotatebox[origin=c]{90}{FS}}    & \multicolumn{2}{l}{ChatGPT}     & 45.8\phantom{$^{\pm0.0}$}  & 50.7\phantom{$^{\pm0.0}$}  & 41.3\phantom{$^{\pm0.0}$}  & 42.4\phantom{$^{\pm0.0}$}  & 43.5\phantom{$^{\pm0.0}$}  & 38.8\phantom{$^{\pm0.0}$}  & 45.5\phantom{$^{\pm0.0}$}  & 45.5\phantom{$^{\pm0.0}$}  & 45.5\phantom{$^{\pm0.0}$}  & 45.5\phantom{$^{\pm0.0}$}  & 45.5\phantom{$^{\pm0.0}$}  & 45.5\phantom{$^{\pm0.0}$}  \\
                                   &                       &                                                        & \multicolumn{2}{l}{Orca~2~13B}  & 35.1\phantom{$^{\pm0.0}$}  & 41.3\phantom{$^{\pm0.0}$}  & 31.3\phantom{$^{\pm0.0}$}  & 31.5\phantom{$^{\pm0.0}$}  & 30.8\phantom{$^{\pm0.0}$}  & 27.3\phantom{$^{\pm0.0}$}  & 44.8\phantom{$^{\pm0.0}$}  & 44.8\phantom{$^{\pm0.0}$}  & 44.8\phantom{$^{\pm0.0}$}  & 44.8\phantom{$^{\pm0.0}$}  & 44.8\phantom{$^{\pm0.0}$}  & 44.8\phantom{$^{\pm0.0}$}  \\                                             
                                                         \cdashlinelr{3-17}
                                   &                       & \multirow{7}{*}{\rotatebox[origin=c]{90}{Fine-tuning}} 
                                                         & \multicolumn{2}{l}{Orca~2~13B}  & 72.6$^{\pm0.6}$            & 68.5$^{\pm0.9}$            & 62.9$^{\pm3.5}$            & 70.2$^{\pm0.5}$            & 70.4$^{\pm0.2}$            & 54.0$^{\pm0.4}$            & 80.7$^{\pm1.0}$            & 80.7$^{\pm1.0}$            & 80.7$^{\pm1.0}$            & 80.7$^{\pm1.0}$            & 80.7$^{\pm1.0}$            & 80.7$^{\pm1.0}$            \\
                                                         \cdashlinelr{4-17}
                                    &                      &                                                        & \multirow{4}{*}{mT5} & Without CD   & 73.5$^{\pm0.8}$            & 70.4$^{\pm0.7}$            & 63.7$^{\pm0.8}$            & 68.8$^{\pm0.5}$            & 73.2$^{\pm0.5}$            & 59.1$^{\pm0.5}$            & 75.7$^{\pm0.4}$            & 75.7$^{\pm0.4}$            & 75.7$^{\pm0.4}$            & 75.7$^{\pm0.4}$            & 75.7$^{\pm0.4}$            & 75.7$^{\pm0.4}$            \\
                                    &                      &                                                        &                      & With CD  & 73.6$^{\pm0.5}$            & 69.9$^{\pm0.4}$            & 64.9$^{\pm0.5}$            & 62.9$^{\pm0.5}$            & 72.8$^{\pm1.0}$            & 60.4$^{\pm2.1}$            & 75.7$^{\pm1.3}$            & 75.7$^{\pm1.3}$            & 75.7$^{\pm1.3}$            & 75.7$^{\pm1.3}$            & 75.7$^{\pm1.3}$            & 75.7$^{\pm1.3}$            \\ 
                                     &                     &                                                        &                      & Without* CD  &
                                                         73.1$^{\pm0.5}$  & 70.1$^{\pm0.8}$ & 64.7$^{\pm0.4}$ & 69.3$^{\pm1.1}$ & 73.0$^{\pm0.4}$ & 62.9$^{\pm1.7}$ & 75.9$^{\pm0.7}$ & 75.9$^{\pm0.7}$ & 75.9$^{\pm0.7}$ & 75.9$^{\pm0.7}$ & 75.9$^{\pm0.7}$ & 75.9$^{\pm0.7}$\\
                                    &                      
                                                         &                                                        &                      & With CD  
                                                         & 73.7$^{\pm0.7}$              & 69.7$^{\pm0.7}$ & 64.5$^{\pm0.8}$ & 64.0$^{\pm1.9}$             & 73.2$^{\pm0.6}$             & 60.7$^{\pm0.8}$ & 76.1$^{\pm0.9}$ & 76.1$^{\pm0.9}$ & 76.1$^{\pm0.9}$ & 76.1$^{\pm0.9}$ & 76.1$^{\pm0.9}$ & 76.1$^{\pm0.9}$ \\
                                                         \cdashlinelr{4-17}
                                    &                      &                                                        & \multirow{2}{*}{mBART} & Without CD   
                                                         & 70.1$^{\pm0.8}$ & 66.4$^{\pm1.6}$ & 61.1$^{\pm1.6}$ & 64.1$^{\pm1.2}$ & 70.9$^{\pm0.6}$ & 56.8$^{\pm2.2}$ & 71.6$^{\pm1.3}$ & 71.6$^{\pm1.3}$ & 71.6$^{\pm1.3}$ & 71.6$^{\pm1.3}$ & 71.6$^{\pm1.3}$ & 71.6$^{\pm1.3}$\\
                                    &                      &                                                        &  & Without CD
                                                         & 68.4$^{\pm0.9}$ & 66.8$^{\pm1.5}$ & 58.2$^{\pm1.2}$ & 58.0$^{\pm1.2}$ & 67.4$^{\pm0.3}$ & 55.3$^{\pm1.5}$ & 70.3$^{\pm0.7}$ & 70.3$^{\pm0.7}$ & 70.3$^{\pm0.7}$ & 70.3$^{\pm0.7}$ & 70.3$^{\pm0.7}$ & 70.3$^{\pm0.7}$\\
                                                         
                                                         \cmidrule{2-17}
& \multirow{7}{*}{\rotatebox[origin=c]{90}{Cross-lingual}} & \multirow{7}{*}{\rotatebox[origin=c]{90}{Fine-tuning}}                                      
                                                         & \multicolumn{2}{l}{Orca~2~13B}  & 54.6$^{\pm0.4}$            & 62.6$^{\pm1.2}$            & 57.1$^{\pm2.4}$            & 53.4$^{\pm3.8}$            & 59.3$^{\pm1.1}$            & 38.0$^{\pm2.2}$            & \underline{73.8}$^{\pm0.2}$            & \underline{68.9}$^{\pm0.1}$            & \underline{71.8}$^{\pm0.1}$            & 63.9$^{\pm0.7}$            & \underline{73.1}$^{\pm0.4}$            & \underline{63.3}$^{\pm0.2}$            \\
                                                         \cdashlinelr{4-17}
                                    &                      &                                                        & \multirow{4}{*}{mT5} & Without CD   & 54.3$^{\pm1.6}$            & 52.5$^{\pm1.0}$            & 55.8$^{\pm0.7}$            & 52.3$^{\pm1.3}$            & 55.0$^{\pm2.7}$            & 41.4$^{\pm1.4}$            & 55.8$^{\pm2.0}$            & 53.6$^{\pm1.7}$            & 61.2$^{\pm1.3}$            & 57.6$^{\pm2.4}$            & 52.8$^{\pm1.0}$            & 44.2$^{\pm1.2}$            \\
                                    &                      &                                                        &                      & With CD  & \textbf{58.7}$^{\pm1.0}$   & \textbf{62.8}$^{\pm1.4}$   & \textbf{57.5}$^{\pm0.3}$   & \textbf{\underline{54.1}}$^{\pm0.2}$   & \textbf{60.4}$^{\pm0.9}$   & \textbf{\underline{49.0}}$^{\pm0.9}$   & \textbf{66.1}$^{\pm1.3}$   & \textbf{65.3}$^{\pm1.2}$   & \textbf{65.1}$^{\pm0.8}$   & \textbf{\underline{65.5}}$^{\pm0.7}$   & \textbf{65.9}$^{\pm0.7}$   & \textbf{55.8}$^{\pm0.6}$   \\
                                     &                     &                                                        &                      & Without CD*  & 51.1$^{\pm1.1}$            & 50.2$^{\pm0.3}$            & 52.5$^{\pm1.5}$            & 47.4$^{\pm1.4}$            & 37.9$^{\pm1.1}$            & 33.7$^{\pm1.2}$            & 41.5$^{\pm2.1}$            & 42.2$^{\pm1.4}$            & 54.3$^{\pm3.1}$            & 50.0$^{\pm1.0}$            & 31.9$^{\pm0.9}$            & 39.6$^{\pm1.3}$            \\
                                    &                      &                                                        &                      & With CD* & \textbf{\underline{59.5}}$^{\pm1.4}$   & \textbf{\underline{63.3}}$^{\pm1.0}$   & \textbf{\underline{58.2}}$^{\pm0.7}$   & \textbf{53.8}$^{\pm0.9}$   & \textbf{\underline{60.6}}$^{\pm1.1}$   & \textbf{44.8}$^{\pm1.6}$   & \textbf{60.5}$^{\pm1.0}$   & \textbf{61.7}$^{\pm1.2}$   & \textbf{63.3}$^{\pm1.1}$   & \textbf{65.0}$^{\pm0.5}$   & \textbf{53.9}$^{\pm2.5}$   & \textbf{56.4}$^{\pm1.5}$   \\ 
                                                         \cdashlinelr{4-17}
                                    &                      &                                                        & \multirow{2}{*}{mBART} & Without CD   
                                                          & 48.6$^{\pm2.7}$   & 52.5$^{\pm1.4}$   & 49.3$^{\pm1.5}$   & 44.5$^{\pm1.4}$   & 53.8$^{\pm1.5}$   & 31.1$^{\pm2.1}$   & 58.4$^{\pm3.5}$ & 59.4$^{\pm4.1}$ & 62.0$^{\pm3.5}$ & 56.6$^{\pm1.9}$ & 56.7$^{\pm3.4}$ & 50.9$^{\pm3.8}$ \\
                                                          
                                    &                       &                                                        & & With CD &
                                                          45.8$^{\pm4.5}$          & \textbf{54.8}$^{\pm0.4}$ & 49.2$^{\pm0.6}$          & \textbf{46.9}$^{\pm0.9}$ & \textbf{55.9}$^{\pm0.2}$ & \textbf{34.7}$^{\pm1.1}$                                                           & 61.7$^{\pm2.1}$          & 59.4$^{\pm2.3}$          & 59.5$^{\pm0.7}$          & \textbf{59.6}$^{\pm1.0}$ & \textbf{61.3}$^{\pm1.1}$ & \textbf{55.7}$^{\pm0.7}$ \\

                                                         \bottomrule
\end{tabular}
\end{adjustbox}
\label{tab:res_pair}
\end{table}

In monolingual settings, ChatGPT consistently achieves the best results in both zero-shot (ZS) and few-shot (FS) configurations. The superior performance of ChatGPT can be attributed to its extensive parameter count and comprehensive training data. Few-shot learning improves performance across all models and tasks. Fine-tuning offers the significant gains, as shown for Orca~2~13B. Among sequence-to-sequence models, mT5 generally performs well despite its smaller size compared to LLMs, outperforming Orca~2 in certain lower-resource languages like Czech and Turkish. The mBART model performs consistently worse than mT5 across all tasks and languages. Constrained decoding does not affect monolingual results.

In the cross-lingual setting, mT5 generally outperforms large language models when English is the source language. All evaluated LLMs perform particularly poorly when Turkish is the target language. However, when English is the target language, Orca~2 often performs best. The difference in performance in English may be related to the predominant pre-training of models like Orca~2 on English data, which gives them a significant advantage in that language. Nonetheless, using English as the target language is less practical due to its resource-rich status and is, therefore, more commonly used as the source language in cross-language tasks. Similar to monolingual results, mT5 outperforms mBART in the majority of cases. While mBART also benefits from constrained decoding, it does so to a lesser extent than mT5.

Constrained decoding proves critical for tasks that involve aspect term prediction (E2E-ABSA and ACTE) in cross-lingual settings. It effectively addresses the issue of models generating aspect terms in the source language rather than the target language -- a problem that also occurs in the ATE task. This issue frequently arises in zero-shot cross-lingual transfer and is particularly problematic for multi-tasking models, where constrained decoding often leads to over 10\% absolute improvements in F1 score, bringing their performance on par with task-specific models. 

The ACSA task, in contrast, does not benefit from constrained decoding. Since ACSA does not involve aspect term prediction, the primary source of cross-lingual transfer errors addressed by constrained decoding is absent. This aligns with findings from simpler tasks like ACD (see Section~\ref{sub:res_simple}). Surprisingly, even multi-tasking models see no improvement from constrained decoding in ACSA. Nevertheless, the general performance pattern holds: ChatGPT leads in zero-shot and few-shot settings, and fine-tuned Orca~2~13B performs best in most cases. The mT5 model shows strong results in some languages, particularly Turkish and Czech. ACSA is comparatively simpler due to the absence of aspect terms, as the label space for aspect categories and sentiment polarities is more limited than the open-ended label space for aspect terms.

We improve on previous cross-lingual E2E-ABSA results~\cite{li2020unsupervised,lin2023clxabsa,zhang-etal-2021-cross}, except for the en$\rightarrow$nl combination. However, this comparison is not straightforward. Prior works use encoder-only models, are restricted to one sentiment polarity per aspect term, and do not predict \textit{\quotes{NULL}} (implicit) aspect terms. In contrast, our method supports multiple sentiment polarities per aspect and includes implicit aspect prediction, making the task more complex. This also changes the number of tuples to be predicted: for instance, our English test set contains 859 tuples, compared to 612 in~\cite{zhang-etal-2021-cross}, where implicit aspects are excluded and sentiment polarities merged. Despite these differences, our constrained decoding approach achieves comparable or superior results in zero-shot cross-lingual settings.

Importantly, our approach does not rely on external translation systems. Previous methods depend on translation tools and are subject to translation quality. In contrast, constrained decoding offers a language-agnostic and implementation-friendly solution. Given its simplicity and effectiveness, our method provides a practical alternative for cross-lingual ABSA tasks.

\subsection{TASD Results}
This section focuses on the TASD task in more detail, given that this task offers the most comprehensive analysis of reviews involving the simultaneous prediction of three sentiment elements. Table~\ref{tab:res_tasd} presents the results of the TASD task.

\begin{table}[ht!]
\centering
\caption{Monolingual (for target languages) and cross-lingual F1 scores for the TASD task. \textbf{Bold} indicates significant improvements with constrained decoding (CD) over without. The best cross-lingual result per task and language pair is \underline{underlined}. Asterisks (*) denote multi-tasking models. ZS and FS stand for zero-shot and few-shot (10 examples).}
\begin{adjustbox}{width=\linewidth}
\begin{tabular}{@{}llllcccccccccccc@{}}
\toprule
\multicolumn{2}{c}{\textbf{Setup}}                                                                                 & \textbf{Model}       &          & \textbf{En$\rightarrow$cs} & \textbf{En$\rightarrow$es} & \textbf{En$\rightarrow$fr} & \textbf{En$\rightarrow$nl} & \textbf{En$\rightarrow$ru} & \textbf{En$\rightarrow$tr}  & \textbf{Cs$\rightarrow$en} & \textbf{Es$\rightarrow$en} & \textbf{Fr$\rightarrow$en} & \textbf{Nl$\rightarrow$en} & \textbf{Ru$\rightarrow$en} & \textbf{Tr$\rightarrow$en} \\ \midrule
\multirow{11}{*}{\rotatebox[origin=c]{90}{Monolingual}}  & \multirow{2}{*}{\rotatebox[origin=c]{90}{ZS}}   & \multicolumn{2}{l}{ChatGPT}     & 25.4\phantom{$^{\pm0.0}$}  & 27.5\phantom{$^{\pm0.0}$}  & 25.1\phantom{$^{\pm0.0}$}  & 18.6\phantom{$^{\pm0.0}$}  & 22.2\phantom{$^{\pm0.0}$}  & 19.2\phantom{$^{\pm0.0}$}   & 27.7\phantom{$^{\pm0.0}$}  & 27.7\phantom{$^{\pm0.0}$}  & 27.7\phantom{$^{\pm0.0}$}  & 27.7\phantom{$^{\pm0.0}$}  & 27.7\phantom{$^{\pm0.0}$}  & 27.7\phantom{$^{\pm0.0}$}  \\
                                                         &                                                        & \multicolumn{2}{l}{Orca~2~13B}  & 11.2\phantom{$^{\pm0.0}$}  & 12.8\phantom{$^{\pm0.0}$}  & 10.5\phantom{$^{\pm0.0}$}  & 9.1\phantom{$^{\pm0.0}$}   & 9.3\phantom{$^{\pm0.0}$}   & 4.6\phantom{$^{\pm0.0}$}    & 18.7\phantom{$^{\pm0.0}$}  & 18.7\phantom{$^{\pm0.0}$}  & 18.7\phantom{$^{\pm0.0}$}  & 18.7\phantom{$^{\pm0.0}$}  & 18.7\phantom{$^{\pm0.0}$}  & 18.7\phantom{$^{\pm0.0}$}  \\                       
                                                         \cdashlinelr{2-16}
                                                         & \multirow{2}{*}{\rotatebox[origin=c]{90}{FS}}    & \multicolumn{2}{l}{ChatGPT}     & 42.6\phantom{$^{\pm0.0}$}  & 47.7\phantom{$^{\pm0.0}$}  & 37.1\phantom{$^{\pm0.0}$}  & 40.0\phantom{$^{\pm0.0}$}  & 37.0\phantom{$^{\pm0.0}$}  & 35.8\phantom{$^{\pm0.0}$}   & 42.4\phantom{$^{\pm0.0}$}  & 42.4\phantom{$^{\pm0.0}$}  & 42.4\phantom{$^{\pm0.0}$}  & 42.4\phantom{$^{\pm0.0}$}  & 42.4\phantom{$^{\pm0.0}$}  & 42.4\phantom{$^{\pm0.0}$}  \\
                                                         &                                                        & \multicolumn{2}{l}{Orca~2~13B}  & 32.6\phantom{$^{\pm0.0}$}  & 44.2\phantom{$^{\pm0.0}$}  & 30.0\phantom{$^{\pm0.0}$}  & 26.5\phantom{$^{\pm0.0}$}  & 28.3\phantom{$^{\pm0.0}$}  & 24.2\phantom{$^{\pm0.0}$}   & 46.3\phantom{$^{\pm0.0}$}  & 46.3\phantom{$^{\pm0.0}$}  & 46.3\phantom{$^{\pm0.0}$}  & 46.3\phantom{$^{\pm0.0}$}  & 46.3\phantom{$^{\pm0.0}$}  & 46.3\phantom{$^{\pm0.0}$}  \\
                                                         \cdashlinelr{2-16}
                                                         & \multirow{7}{*}{\rotatebox[origin=c]{90}{Fine-tuning}}
                                                         & \multicolumn{2}{l}{Orca~2~13B}  & 65.6$^{\pm0.4}$            & 66.2$^{\pm0.4}$            & 63.0$^{\pm1.0}$            & 64.5$^{\pm1.8}$            & 64.5$^{\pm1.0}$            & 48.8$^{\pm0.9}$             & 77.3$^{\pm1.8}$            & 77.3$^{\pm1.8}$            & 77.3$^{\pm1.8}$            & 77.3$^{\pm1.8}$            & 77.3$^{\pm1.8}$            & 77.3$^{\pm1.8}$            \\
                                                         \cdashlinelr{3-16}
                                                         &                                                        & \multirow{4}{*}{mT5} & Without CD   & 66.9$^{\pm0.3}$            & 65.8$^{\pm0.4}$            & 59.0$^{\pm0.6}$            & 62.9$^{\pm1.4}$            & 67.0$^{\pm0.9}$            & 54.1$^{\pm3.0}$             & 71.4$^{\pm0.9}$            & 71.4$^{\pm0.9}$            & 71.4$^{\pm0.9}$            & 71.4$^{\pm0.9}$            & 71.4$^{\pm0.9}$            & 71.4$^{\pm0.9}$            \\
                                                         &                                                        &                      & With CD  & 67.1$^{\pm1.3}$            & 66.2$^{\pm0.5}$            & 58.9$^{\pm1.1}$            & 57.6$^{\pm0.5}$            & 66.4$^{\pm0.4}$            & 53.9$^{\pm1.5}$             & 70.4$^{\pm0.8}$            & 70.4$^{\pm0.8}$            & 70.4$^{\pm0.8}$            & 70.4$^{\pm0.8}$            & 70.4$^{\pm0.8}$            & 70.4$^{\pm0.8}$            \\
                                                         
                                                         &                                                        &                      & Without CD*  &
                                                         66.3$^{\pm0.5}$ & 65.6$^{\pm0.5}$ & 57.9$^{\pm0.6}$ & 62.8$^{\pm0.9}$ & 65.7$^{\pm0.8}$ & 58.0$^{\pm0.8}$ & 70.5$^{\pm0.3}$ & 70.5$^{\pm0.3}$ & 70.5$^{\pm0.3}$ & 70.5$^{\pm0.3}$ & 70.5$^{\pm0.3}$ & 70.5$^{\pm0.3}$\\
                                                         
                                                         &                                                        &                      & With CD*  
                                                         & 67.3$^{\pm0.5}$           & 64.6$^{\pm0.4}$ & 58.0$^{\pm0.8}$ & 58.0$^{\pm1.7}$             & 66.5$^{\pm0.4}$             & 57.0$^{\pm0.5}$   & 70.8$^{\pm0.8}$   & 70.8$^{\pm0.8}$   & 70.8$^{\pm0.8}$   & 70.8$^{\pm0.8}$   & 70.8$^{\pm0.8}$   & 70.8$^{\pm0.8}$\\
                                                         \cdashlinelr{3-16}
                                                         &                                                        & \multirow{2}{*}{mBART} & Without CD   
                                                         & 62.6$^{\pm0.7}$ & 62.9$^{\pm1.2}$ & 54.8$^{\pm0.9}$ & 57.6$^{\pm0.9}$ & 62.6$^{\pm0.7}$ & 49.3$^{\pm3.1}$ & 66.1$^{\pm1.4}$ & 66.1$^{\pm1.4}$ & 66.1$^{\pm1.4}$ & 66.1$^{\pm1.4}$ & 66.1$^{\pm1.4}$ & 66.1$^{\pm1.4}$\\
                                                         &                                                        &  & With CD   
                                                         & 61.9$^{\pm1.6}$ & 61.5$^{\pm1.4}$ & 52.4$^{\pm0.6}$ & 52.1$^{\pm1.0}$ & 60.1$^{\pm1.9}$ & 47.6$^{\pm2.7}$ & 64.8$^{\pm1.4}$ & 64.8$^{\pm1.4}$ & 64.8$^{\pm1.4}$ & 64.8$^{\pm1.4}$ & 64.8$^{\pm1.4}$ & 64.8$^{\pm1.4}$\\
                                                         
                                                         \midrule
\multirow{7}{*}{\rotatebox[origin=c]{90}{Cross-lingual}} & \multirow{7}{*}{\rotatebox[origin=c]{90}{Fine-tuning}}                                      
                                                         & \multicolumn{2}{l}{Orca~2~13B}  & 49.7$^{\pm0.5}$            & \underline{58.0}$^{\pm1.1}$            & \underline{56.1}$^{\pm1.0}$            & 50.2$^{\pm1.1}$            & \underline{55.6}$^{\pm1.4}$            & 31.4$^{\pm1.6}$             & \underline{71.7}$^{\pm0.2}$            & \underline{67.2}$^{\pm0.3}$            & \underline{68.3}$^{\pm1.6}$            & 59.1$^{\pm1.4}$            & \underline{69.2}$^{\pm0.6}$            & \underline{60.4}$^{\pm0.1}$            \\
                                                         \cdashlinelr{3-16}
                                                         &                                                        & \multirow{4}{*}{mT5} & Without CD   & 50.2$^{\pm0.9}$            & 48.3$^{\pm0.5}$            & 50.4$^{\pm1.4}$            & 47.7$^{\pm1.1}$            & 48.6$^{\pm2.0}$            & 39.1$^{\pm3.6}$             & 54.9$^{\pm1.9}$            & 50.7$^{\pm1.7}$            & 57.4$^{\pm1.0}$            & 53.3$^{\pm0.8}$            & 50.1$^{\pm2.0}$            & 43.2$^{\pm0.6}$            \\
                                                         &                                                        &                      & With CD  & \textbf{53.3}$^{\pm1.5}$   & \textbf{57.6}$^{\pm0.6}$   & 50.4$^{\pm0.8}$            & \textbf{\underline{50.4}}$^{\pm1.3}$   & \textbf{54.9}$^{\pm2.0}$   & \textbf{\underline{43.8}}$^{\pm0.8}$    & \textbf{60.1}$^{\pm0.8}$   & \textbf{59.7}$^{\pm0.4}$   & \textbf{59.9}$^{\pm0.6}$   & \textbf{\underline{59.9}}$^{\pm0.3}$   & \textbf{61.8}$^{\pm0.8}$   & \textbf{52.3}$^{\pm0.5}$   \\
                                                          &                                                        &                      & Without CD*  & 45.4$^{\pm1.5}$            & 46.2$^{\pm0.7}$            & 44.8$^{\pm1.6}$            & 43.4$^{\pm0.8}$            & 33.6$^{\pm1.3}$            & 24.7$^{\pm4.0}$            & 36.5$^{\pm2.1}$            & 37.7$^{\pm0.9}$            & 49.8$^{\pm2.3}$            & 44.7$^{\pm0.9}$            & 28.7$^{\pm0.7}$            & 35.3$^{\pm0.8}$            \\
                                                       &                                                        &                      & With CD*    & \textbf{\underline{53.4}}$^{\pm1.5}$   & \textbf{57.8}$^{\pm0.5}$    & \textbf{50.8}$^{\pm1.1}$    & \textbf{49.3}$^{\pm0.6}$             & \textbf{53.6}$^{\pm0.9}$    & \textbf{40.2}$^{\pm1.6}$             & \textbf{55.7}$^{\pm1.3}$    & \textbf{56.1}$^{\pm0.6}$    & \textbf{58.0}$^{\pm0.8}$    & \textbf{58.0}$^{\pm1.1}$             & \textbf{50.2}$^{\pm2.7}$    & \textbf{50.5}$^{\pm1.3}$    \\ 
                                                       \cdashlinelr{3-16}
                                                       &                                                        & \multirow{2}{*}{mBART} & Without CD
                                                       & 40.4$^{\pm3.0}$   & 47.6$^{\pm1.9}$   & 39.6$^{\pm0.8}$   & 39.1$^{\pm0.9}$   & 48.5$^{\pm1.1}$   & 23.5$^{\pm2.6}$ & 54.3$^{\pm0.8}$ & 55.4$^{\pm2.9}$ & 56.0$^{\pm2.3}$ & 50.4$^{\pm1.4}$ & 54.5$^{\pm3.2}$ & 49.1$^{\pm0.8}$  \\   &                                                        &  & With CD& 39.3$^{\pm1.0}$          & \textbf{51.1}$^{\pm1.2}$ & 39.9$^{\pm0.6}$          & 38.9$^{\pm0.9}$          & \textbf{50.5}$^{\pm0.7}$ & \textbf{27.3}$^{\pm1.1}$  & 54.2$^{\pm1.7}$          & 55.1$^{\pm1.9}$          & 55.0$^{\pm1.3}$          & \textbf{54.4}$^{\pm1.1}$ & \textbf{59.0}$^{\pm0.9}$ & 48.8$^{\pm0.8}$ \\
                                                          \bottomrule

\end{tabular}
\end{adjustbox}
\label{tab:res_tasd}
\end{table}

The monolingual results for the TASD task show that ChatGPT consistently performs the best across different settings. However, zero-shot performance is generally weak, with F1 scores often falling below 10\% for non-English languages for models other than ChatGPT, and under 20\% for English. Even ChatGPT does not exceed 30\% for any language in zero-shot scenarios. Few-shot settings yield better results but remain under 50\% for all languages while often not reaching even 30\%, highlighting the inherent difficulty of the TASD task. Fine-tuning improves performance significantly, with fine-tuned versions of Orca~2 and mT5 achieving the best monolingual results.

In cross-lingual settings, similar patterns emerge as observed in other tasks. Constrained decoding significantly enhances performance. This improvement is particularly pronounced because the task involves predicting aspect terms, which may be generated in the source language rather than the target language, an issue mitigated by constrained decoding. For multi-tasking models, constrained decoding provides even more substantial benefits. Generally, multi-tasking models perform on par with specialized models. Orca~2 usually achieves the best results when English is the target language. When English is the source language, both mT5 and Orca~2 perform well, except for Turkish as the target language, where mT5 outperforms Orca~2 by 12\%. Compared to monolingual results, cross-lingual results are often around 10\% worse.

Table~\ref{tab:res_tasd_all} shows the results for the TASD task, comparing the performance of the mT5 model with and without constrained decoding alongside the Orca~2~13B model across all language combinations. The results indicate that constrained decoding generally enhances performance across most language pairs. On average, constrained decoding improves the results by almost 5\%. Notably, the mT5 model consistently outperforms the Orca~2~13B model for some target languages, especially for Turkish and Czech. In contrast, the Orca~2 model demonstrates superiority when French and English are the target languages. On average, the Orca~2 model outperforms the mT5 model with constrained decoding by 0.6\% (0.4\% in cross-lingual settings), which is primarily due to English as the target language, where it outperforms the mT5 model by more than 7\% on average. Notably, the mT5 model has over ten times fewer parameters, yet its performance is nearly equivalent to that of the Orca~2 model. Overall, English emerges as the most favourable target language overall, yielding the highest scores across multiple source languages. On the other hand, Turkish consistently yields weaker results, suggesting potential challenges due to limited data availability or greater linguistic divergence. Turkish is the only language in the study that does not come from an Indo-European family.

\begin{table}[ht!]
\centering
\caption{Comparison of mT5 models with and without constrained decoding (CD), and Orca~2~13B across various language combinations for the TASD task. Results are reported in terms of F1 scores, with target languages in columns and source languages in rows. AVG* excludes monolingual results, i.e. results where the source and target languages are the same.}
\begin{adjustbox}{width=0.75\linewidth}

\begin{tabular}{@{}cllccccccccc@{}}
\toprule
                               &                      &            & \textbf{Cs} & \textbf{En} & \textbf{Es} & \textbf{Fr} & \textbf{Nl} & \textbf{Ru} & \textbf{Tr} & \textbf{AVG} & \textbf{AVG*} \\ \midrule
\multirow{3}{*}{\textbf{Cs}}   & \multirow{2}{*}{mT5} & Without CD & 66.9        & 54.9        & 50.1        & 41.9        & 45.1        & 44.4        & 39.2        & 48.9         & 45.9          \\
                               &                      & With CD    & 67.1        & 60.1        & 57.4        & 46.9        & 46.0        & 55.1        & 41.8        & 53.5         & 51.2          \\
                               & \multicolumn{2}{l}{Orca~2~13B}    & 65.6        & 71.7        & 60.7        & 50.8        & 46.3        & 55.5        & 38.4        & 55.6         & 53.9          \\\cdashlinelr{1-12}
\multirow{3}{*}{\textbf{En}}   & \multirow{2}{*}{mT5} & Without CD & 50.2        & 71.4        & 48.3        & 50.4        & 47.7        & 48.6        & 39.1        & 50.8         & 47.4          \\
                               &                      & With CD    & 53.3        & 70.4        & 57.6        & 50.4        & 50.4        & 54.9        & 43.8        & 54.4         & 51.7          \\
                               & \multicolumn{2}{l}{Orca~2~13B}    & 49.7        & 77.3        & 58.0        & 56.1        & 50.2        & 55.6        & 31.4        & 54.0         & 50.1          \\\cdashlinelr{1-12}
\multirow{3}{*}{\textbf{Es}}   & \multirow{2}{*}{mT5} & Without CD & 50.8        & 50.7        & 65.8        & 43.2        & 43.1        & 49.8        & 38.5        & 48.9         & 46.0          \\
                               &                      & With CD    & 55.5        & 59.7        & 66.2        & 49.9        & 45.1        & 52.1        & 43.3        & 53.1         & 50.9          \\
                               & \multicolumn{2}{l}{Orca~2~13B}    & 52.1        & 67.2        & 66.2        & 54.2        & 49.0        & 51.4        & 30.9        & 53.0         & 50.8          \\\cdashlinelr{1-12}
\multirow{3}{*}{\textbf{Fr}}   & \multirow{2}{*}{mT5} & Without CD & 47.8        & 57.4        & 46.6        & 59.0        & 47.4        & 46.1        & 32.5        & 48.1         & 46.3          \\
                               &                      & With CD    & 53.3        & 59.9        & 57.6        & 58.9        & 48.3        & 50.5        & 35.1        & 52.0         & 50.8          \\
                               & \multicolumn{2}{l}{Orca~2~13B}    & 51.5        & 68.3        & 60.1        & 63.0        & 50.2        & 52.3        & 30.7        & 53.7         & 52.2          \\\cdashlinelr{1-12}
\multirow{3}{*}{\textbf{Nl}}   & \multirow{2}{*}{mT5} & Without CD & 43.7        & 53.3        & 48.8        & 43.2        & 60.3        & 43.6        & 29.0        & 46.0         & 43.6          \\
                               &                      & With CD    & 49.1        & 59.9        & 57.1        & 48.0        & 58.5        & 46.1        & 34.5        & 50.5         & 49.1          \\
                               & \multicolumn{2}{l}{Orca~2~13B}    & 42.4        & 59.1        & 53.2        & 49.1        & 63.3        & 41.1        & 25.1        & 47.6         & 45.0          \\\cdashlinelr{1-12}
\multirow{3}{*}{\textbf{Ru}}   & \multirow{2}{*}{mT5} & Without CD & 41.6        & 51.3        & 52.5        & 44.5        & 41.6        & 67.0        & 37.0        & 47.9         & 44.8          \\
                               &                      & With CD    & 53.8        & 59.4        & 55.4        & 46.8        & 47.4        & 66.4        & 40.4        & 52.8         & 50.5          \\
                               & \multicolumn{2}{l}{Orca~2~13B}    & 54.7        & 67.7        & 57.0        & 50.9        & 49.2        & 64.5        & 29.8        & 53.4         & 51.5          \\\cdashlinelr{1-12}
\multirow{3}{*}{\textbf{Tr}}   & \multirow{2}{*}{mT5} & Without CD & 45.2        & 43.2        & 37.4        & 30.7        & 36.1        & 42.4        & 54.1        & 41.3         & 39.2          \\
                               &                      & With CD    & 47.4        & 52.3        & 40.9        & 32.2        & 39.1        & 43.7        & 53.9        & 44.2         & 42.6          \\
                               & \multicolumn{2}{l}{Orca~2~13B}    & 49.2        & 60.4        & 50.8        & 36.4        & 36.0        & 47.4        & 48.8        & 47.0         & 46.7          \\\midrule
\multirow{3}{*}{\textbf{AVG}}  & \multirow{2}{*}{mT5} & Without CD & 49.5        & 54.6        & 49.9        & 44.7        & 45.9        & 48.9        & 38.5        & 47.4         & --             \\
                               &                      & With CD    & 54.2        & 60.2        & 56.0        & 47.6        & 47.8        & 52.7        & 41.8        & 51.5         & --             \\
                               & \multicolumn{2}{l}{Orca~2~13B}    & 52.2        & 67.4        & 58.0        & 51.5        & 49.2        & 52.5        & 33.6        & 52.1         & --             \\\cdashlinelr{1-12}
\multirow{3}{*}{\textbf{AVG*}} & \multirow{2}{*}{mT5} & Without CD & 46.6        & 51.8        & 47.3        & 42.3        & 43.5        & 45.8        & 35.9        & --            & 44.7          \\
                               &                      & With CD    & 52.1        & 58.6        & 54.3        & 45.7        & 46.1        & 50.4        & 39.8        & --            & 49.6          \\
                               & \multicolumn{2}{l}{Orca~2~13B}    & 49.9        & 65.7        & 56.6        & 49.6        & 46.8        & 50.6        & 31.1        & --            & 50.0         \\\bottomrule
\end{tabular}
\end{adjustbox}
\label{tab:res_tasd_all}
\end{table}

\subsection{Detailed LLM Results}

Table~\ref{tab:res_llm_mono} shows the monolingual results for LLMs. The best results are generally achieved with the Orca~2~13B and LLaMA~3~8B models in zero-shot and few-shot settings. The LLaMA~3 is a more modern model, which may contribute to its good performance despite having fewer parameters than the LLaMA~2~13B and Orca~2~13B. Orca~2 is based on the LLaMA~2 with some improvements, which could explain why Orca~2 models outperform the same-sized LLaMA~2 models. The Orca~2~13B model generally achieves the best results with fine-tuning. Interestingly, the Orca~2~7B model often outperforms the larger LLaMA~2~13B model with fine-tuning, suggesting that smaller but more advanced models can outperform larger, less sophisticated ones.

\begin{table}[ht!]
\centering
\caption{Zero-shot, few-shot and monolingual results for compound ABSA tasks with five different LLMs. The best zero-shot, few-shot and fine-tuning results for each combination of task and language are in \textbf{bold}.}
\begin{adjustbox}{width=\linewidth}
\begin{tabular}{@{}lllccccccc@{}}
\toprule
\textbf{Model}                                       & \textbf{Settings}                         & \textbf{Task} & \textbf{En}                   & \textbf{Cs}                   & \textbf{Es}                   & \textbf{Fr}                   & \textbf{Nl}          & \textbf{Ru}                   & \textbf{Tr}                   \\ \midrule
\multirow{8}{*}{LLaMA~2~13B}      & \multirow{4}{*}{Zero- / few-shot} & ACSA          & 43.3 / 60.6                   & 32.7 / 49.0                   & 37.8 / 60.2                   & 34.0 / 43.3                   & 36.5 / 52.3          & 35.5 / 54.8                   & 24.5 / 34.9                   \\
                                                     &                                   & E2E           & 34.7 / 46.5                   & 19.0 / \textbf{44.0}          & 26.2 / 52.3                   & 24.2 / 38.2                   & 18.7 / 39.7          & 13.9 / 34.0                   & 13.8 / 23.3                   \\
                                                     &                                   & ACTE          & 17.9 / 34.7                   & 9.6 / 37.7                    & 8.6 / 42.7                    & 8.8 / 32.1                    & 6.8 / 25.0           & 5.4 / 29.4                    & 4.4 / 24.0                    \\
                                                     &                                   & TASD          & 17.5 / 33.1                   & 10.2 / \textbf{35.7}          & 11.1 / 34.6                   & 8.4 / 24.1                    & 8.0 / 23.9           & 6.1 / 24.6                    & \textbf{6.9} / 18.7           \\
                                                     \cdashlinelr{2-10}
                                                     & \multirow{4}{*}{Fine-tuning}      & ACSA          & 82.5$^{\pm0.9}$               & 73.2$^{\pm1.7}$               & 78.4$^{\pm1.3}$               & 69.3$^{\pm3.2}$               & 74.7$^{\pm1.7}$      & 77.2$^{\pm1.7}$               & 67.0$^{\pm2.2}$               \\
                                                     &                                   & E2E           & 77.8$^{\pm4.1}$               & 69.4$^{\pm0.7}$               & 69.4$^{\pm1.2}$               & 65.8$^{\pm0.7}$               & 71.3$^{\pm1.8}$      & 67.3$^{\pm1.4}$               & 48.0$^{\pm2.8}$               \\
                                                     &                                   & ACTE          & 75.6$^{\pm1.2}$               & 67.0$^{\pm0.9}$               & 65.4$^{\pm3.1}$               & 58.2$^{\pm5.1}$               & 66.0$^{\pm1.3}$      & 68.8$^{\pm1.3}$               & 53.7$^{\pm1.8}$               \\
                                                     &                                   & TASD          & 72.0$^{\pm0.8}$               & 60.7$^{\pm0.9}$               & 61.6$^{\pm1.8}$               & 59.3$^{\pm0.8}$               & 60.3$^{\pm0.9}$      & 62.1$^{\pm1.8}$               & 41.6$^{\pm1.6}$               \\
                                                     \midrule
\multirow{8}{*}{LLaMA~2~7B}       & \multirow{4}{*}{Zero- / few-shot} & ACSA          & 13.3 / 57.3                   & 14.1 / 41.0                   & 17.8 / 50.7                   & 14.1 / 42.8                   & 15.3 / 14.4          & 14.4 / 37.0                   & 19.5 / 38.7                   \\
                                                     &                                   & E2E           & 24.4 / 40.9                   & 12.2 / 35.0                   & 15.2 / 42.9                   & 13.8 / 32.4                   & 10.7 / 4.1           & 6.5 / 26.2                    & 11.4 / 18.7                   \\
                                                     &                                   & ACTE          & 11.7 / 35.1                   & 6.0 / 29.1                    & 5.6 / 30.0                    & 5.8 / 28.0                    & 2.7 / 10.1           & 1.7 / 14.3                    & 4.3 / 17.5                    \\
                                                     &                                   & TASD          & 7.9 / 30.6                    & 3.6 / 25.9                    & 4.8 / 23.5                    & 3.1 / 13.9                    & 2.0 / 15.2           & 1.0 / 11.4                    & 2.7 / 11.7                    \\
                                                     \cdashlinelr{2-10}
                                                     & \multirow{4}{*}{Fine-tuning}      & ACSA          & 82.1$^{\pm1.2}$               & 73.3$^{\pm0.7}$               & 74.3$^{\pm3.8}$               & 66.6$^{\pm0.7}$               & 71.4$^{\pm1.0}$      & 74.8$^{\pm3.1}$               & 61.0$^{\pm3.3}$               \\
                                                     &                                   & E2E           & 75.5$^{\pm2.4}$               & 66.5$^{\pm0.4}$               & 67.8$^{\pm3.9}$               & 60.6$^{\pm0.6}$               & 64.8$^{\pm1.2}$      & 65.3$^{\pm1.5}$               & 44.4$^{\pm1.1}$               \\
                                                     &                                   & ACTE          & 73.4$^{\pm1.1}$               & 66.1$^{\pm0.7}$               & 65.3$^{\pm0.9}$               & 56.4$^{\pm4.4}$               & 62.7$^{\pm1.7}$      & 66.7$^{\pm1.2}$               & 46.9$^{\pm1.8}$               \\
                                                     &                                   & TASD          & 70.3$^{\pm1.7}$               & 60.2$^{\pm0.8}$               & 58.1$^{\pm4.5}$               & 54.0$^{\pm0.7}$               & 55.5$^{\pm0.3}$      & 58.8$^{\pm1.0}$               & 36.9$^{\pm1.9}$               \\
                                                     \midrule
\multirow{8}{*}{LLaMA~3~8B} & \multirow{4}{*}{Zero- / few-shot} & ACSA          & 58.9 / 62.5                   & \textbf{53.3} / 59.0          & \textbf{54.5} / \textbf{67.2} & \textbf{51.3} / 37.7          & \textbf{48.7} / 57.7 & 50.5 / \textbf{60.3}          & \textbf{44.7} / \textbf{57.3} \\
                                                     &                                   & E2E           & 40.9 / 55.7                   & \textbf{34.5} / 42.3          & \textbf{40.6} / \textbf{55.5} & \textbf{27.5} / 41.4          & \textbf{30.9} / 40.9 & \textbf{28.8} / \textbf{42.0} & \textbf{20.0} / \textbf{45.5} \\
                                                     &                                   & ACTE          & 18.4 / \textbf{49.8}          & \textbf{16.7} / \textbf{39.4} & \textbf{19.3} / \textbf{50.3} & 12.5 / \textbf{35.2}          & \textbf{13.5} / 31.6 & 7.9 / \textbf{37.1}           & 7.3 / \textbf{42.0}           \\
                                                     &                                   & TASD          & 10.9 / 43.7                   & 7.9 / 34.6                    & \textbf{13.5} / \textbf{49.0} & 4.8 / 27.1                    & 8.3 / \textbf{32.8}  & 3.6 / \textbf{35.5}           & 5.4 / \textbf{38.3}           \\
                                                     \cdashlinelr{2-10}
                                                     & \multirow{4}{*}{Fine-tuning}      & ACSA          & 81.1$^{\pm2.1}$               & 70.8$^{\pm2.1}$               & 73.2$^{\pm0.5}$               & 69.5$^{\pm3.7}$               & 71.5$^{\pm3.5}$      & 71.6$^{\pm2.3}$               & 65.5$^{\pm1.6}$               \\
                                                     &                                   & E2E           & 71.4$^{\pm2.8}$               & 63.0$^{\pm1.1}$               & 70.0$^{\pm2.0}$               & 63.1$^{\pm1.8}$               & 66.0$^{\pm1.2}$      & 60.7$^{\pm2.3}$               & 48.6$^{\pm2.0}$               \\
                                                     &                                   & ACTE          & 69.2$^{\pm1.7}$               & 63.1$^{\pm0.6}$               & 59.8$^{\pm2.9}$               & 54.9$^{\pm1.2}$               & 58.1$^{\pm4.3}$      & 62.1$^{\pm1.7}$               & 46.3$^{\pm2.9}$               \\
                                                     &                                   & TASD          & 62.5$^{\pm2.4}$               & 56.8$^{\pm1.4}$               & 57.2$^{\pm1.7}$               & 48.2$^{\pm2.4}$               & 55.4$^{\pm2.8}$      & 53.7$^{\pm2.9}$               & 39.1$^{\pm3.2}$               \\
                                                     \midrule
\multirow{8}{*}{Orca~2~13B}                & \multirow{4}{*}{Zero- / few-shot} & ACSA          & \textbf{59.6} / \textbf{65.2} & 49.5 / \textbf{63.4}          & 50.4 / 67.0                   & 46.3 / \textbf{55.7}          & 47.6 / \textbf{61.1} & \textbf{51.0} / 59.8          & 33.0 / 45.6                   \\
                                                     &                                   & E2E           & \textbf{42.2} / \textbf{62.4} & 22.7 / 41.0                   & 27.4 / 50.6                   & 22.5 / \textbf{43.9}          & 18.3 / \textbf{46.1} & 20.0 / 38.7                   & 19.2 / 32.0                   \\
                                                     &                                   & ACTE          & \textbf{19.3} / 44.8          & 13.7 / 35.1                   & 15.3 / 41.3                   & \textbf{14.1} / 31.3          & 10.1 / 31.5          & \textbf{12.4} / 30.8          & \textbf{10.4} / 27.3          \\
                                                     &                                   & TASD          & \textbf{18.7} / \textbf{46.3} & \textbf{11.2} / 32.6          & 12.8 / 44.2                   & \textbf{10.5} / \textbf{30.0} & \textbf{9.1} / 26.5  & \textbf{9.3} / 28.3           & 4.6 / 24.2                    \\
                                                     \cdashlinelr{2-10}
                                                     & \multirow{4}{*}{Fine-tuning}      & ACSA          & \textbf{84.3}$^{\pm0.9}$               & \textbf{75.4}$^{\pm0.6}$               & \textbf{80.4}$^{\pm0.4}$               & \textbf{78.1}$^{\pm0.4}$               & \textbf{76.1}$^{\pm0.7}$      & \textbf{80.2}$^{\pm0.6}$               & \textbf{69.3}$^{\pm0.8}$               \\
                                                     &                                   & E2E           & \textbf{82.3}$^{\pm0.6}$               & \textbf{71.8}$^{\pm1.1}$               & \textbf{74.8}$^{\pm1.0}$               & \textbf{69.8}$^{\pm2.3}$               & \textbf{76.2}$^{\pm0.3}$      & \textbf{72.2}$^{\pm1.2}$               & \textbf{52.2}$^{\pm1.6}$               \\
                                                     &                                   & ACTE          & \textbf{80.7}$^{\pm1.0}$               & \textbf{72.6}$^{\pm0.6}$               & 68.5$^{\pm0.9}$               & 62.9$^{\pm3.5}$               & \textbf{70.2}$^{\pm0.5}$      & 70.4$^{\pm0.2}$               & \textbf{54.0}$^{\pm0.4}$               \\
                                                     &                                   & TASD          & \textbf{77.3}$^{\pm1.8}$               & \textbf{65.6}$^{\pm0.4}$               & \textbf{66.2}$^{\pm0.4}$               & \textbf{63.0}$^{\pm1.0}$               & \textbf{64.5}$^{\pm1.8}$      & \textbf{64.5}$^{\pm1.0}$               & \textbf{48.8}$^{\pm0.9}$               \\
                                                     \midrule
\multirow{8}{*}{Orca~2~7B}                 & \multirow{4}{*}{Zero- / few-shot} & ACSA          & 38.8 / 59.7                   & 35.7 / 54.3                   & 30.6 / 59.0                   & 22.9 / 47.9                   & 23.5 / 45.9          & 22.3 / 51.7                   & 18.9 / 32.3                   \\
                                                     &                                   & E2E           & 36.5 / 49.6                   & 16.1 / 33.0                   & 24.4 / 46.3                   & 22.1 / 30.6                   & 13.7 / 32.9          & 10.2 / 26.9                   & 7.4 / 24.5                    \\
                                                     &                                   & ACTE          & 16.6 / 35.2                   & 7.6 / 28.4                    & 9.4 / 36.3                    & 10.6 / 24.8                   & 6.5 / \textbf{35.8}  & 3.5 / 22.8                    & 9.8 / 21.8                    \\
                                                     &                                   & TASD          & 13.7 / 39.5                   & 4.6 / 26.7                    & 7.7 / 38.1                    & 7.4 / 25.1                    & 3.1 / 32.0           & 3.0 / 18.4                    & 2.2 / 16.6                    \\
                                                     \cdashlinelr{2-10}
                                                     & \multirow{4}{*}{Fine-tuning}      & ACSA          & 83.5$^{\pm0.6}$               & 74.3$^{\pm0.4}$               & 77.4$^{\pm0.5}$               & 72.9$^{\pm0.6}$               & 75.9$^{\pm0.8}$      & 78.1$^{\pm0.6}$               & 66.2$^{\pm0.8}$               \\
                                                     &                                   & E2E           & 81.4$^{\pm0.5}$               & 70.8$^{\pm0.2}$               & 74.5$^{\pm0.1}$               & 67.7$^{\pm0.1}$               & 64.1$^{\pm0.3}$      & 69.3$^{\pm0.2}$               & 47.8$^{\pm1.0}$               \\
                                                     &                                   & ACTE          & 79.9$^{\pm1.3}$               & 70.0$^{\pm0.3}$               & \textbf{68.6}$^{\pm0.2}$               & \textbf{65.1}$^{\pm0.2}$               & 65.5$^{\pm0.2}$      & \textbf{70.5}$^{\pm0.1}$               & 51.3$^{\pm0.6}$               \\
                                                     &                                   & TASD          & 74.8$^{\pm0.3}$               & 63.4$^{\pm0.3}$               & 63.4$^{\pm0.3}$               & 58.8$^{\pm3.9}$               & 60.3$^{\pm1.6}$      & 62.1$^{\pm0.3}$               & 42.0$^{\pm0.8}$               \\ \bottomrule 
\end{tabular}
\end{adjustbox}
\label{tab:res_llm_mono}
\end{table}

Despite strong zero-shot and few-shot results with LLaMA~3, its fine-tuned version often underperforms compared to other models. This may be due to suboptimal fine-tuning hyperparameters, which could be less compatible with the newer architecture. Overall, adding few-shot examples substantially improves performance, while the best results are achieved with fine-tuning.

Table~\ref{tab:res_llm_cl} presents the cross-lingual results with LLMs. Similar to the monolingual results with fine-tuning, the Orca~2~13B model performs best in most cases across all tasks and language combinations.

\begin{table}[ht!]
\centering
\caption{F1 scores for compound ABSA tasks with five different LLMs in cross-lingual fine-tuning settings. The best results for each task and language combination is in \textbf{bold}.}
\begin{adjustbox}{width=\linewidth}
\begin{tabular}{@{}llcccccccccccc@{}}
\toprule
\textbf{Model} & \textbf{Task} 
                                 & \textbf{En$\rightarrow$cs} & \textbf{En$\rightarrow$es} & \textbf{En$\rightarrow$fr} & \textbf{En$\rightarrow$nl} & \textbf{En$\rightarrow$ru} & \textbf{En$\rightarrow$tr} & \textbf{Cs$\rightarrow$en} & \textbf{Es$\rightarrow$en} & \textbf{Fr$\rightarrow$en} & \textbf{Nl$\rightarrow$en} & \textbf{Ru$\rightarrow$en} & \textbf{Tr$\rightarrow$en} \\ \midrule
\multirow{4}{*}{LLaMA~2~13B}               & ACSA                           & 66.6$^{\pm2.0}$        & 72.9$^{\pm1.5}$        & 69.8$^{\pm1.2}$        & \textbf{71.8}$^{\pm1.4}$        & 74.0$^{\pm1.2}$        & 52.0$^{\pm3.9}$        & 76.9$^{\pm1.7}$        & 79.7$^{\pm2.4}$        & 77.5$^{\pm3.9}$        & 77.9$^{\pm0.5}$        & 81.0$^{\pm0.6}$        & 78.8$^{\pm1.1}$        \\
                                & E2E                            & 48.0$^{\pm2.8}$        & 58.0$^{\pm0.7}$        & 52.3$^{\pm5.3}$        & 55.4$^{\pm3.7}$        & 53.3$^{\pm0.9}$        & 29.5$^{\pm4.9}$        & 69.0$^{\pm0.8}$        & 62.7$^{\pm1.7}$        & 62.6$^{\pm5.7}$        & 56.9$^{\pm1.0}$        & 68.8$^{\pm1.6}$        & 57.1$^{\pm1.0}$        \\
                                & ACTE                           & 44.5$^{\pm0.8}$        & 52.1$^{\pm2.6}$        & 49.4$^{\pm2.1}$        & 47.7$^{\pm2.0}$        & 50.6$^{\pm3.1}$        & 31.4$^{\pm2.5}$        & 61.5$^{\pm3.8}$        & 59.6$^{\pm1.3}$        & 58.8$^{\pm4.6}$        & 57.5$^{\pm1.4}$        & 54.7$^{\pm1.7}$        & 57.7$^{\pm0.9}$        \\
                                & TASD                           & 40.4$^{\pm1.9}$        & 49.3$^{\pm1.4}$        & 45.3$^{\pm2.2}$        & 41.8$^{\pm2.0}$        & 43.7$^{\pm3.4}$        & 25.0$^{\pm2.6}$        & 62.4$^{\pm0.9}$        & 55.9$^{\pm0.8}$        & 61.6$^{\pm2.8}$        & 51.8$^{\pm0.6}$        & 53.5$^{\pm3.6}$        & 49.8$^{\pm1.7}$        \\ \cdashlinelr{1-14}
\multirow{4}{*}{LLaMA~2~7B}               & ACSA                           & 64.7$^{\pm2.2}$        & 71.6$^{\pm1.3}$        & 68.0$^{\pm1.2}$        & 65.4$^{\pm1.8}$        & 71.6$^{\pm0.6}$        & 46.1$^{\pm2.0}$        & 79.5$^{\pm0.4}$        & 74.3$^{\pm1.4}$        & 73.6$^{\pm1.4}$        & 77.5$^{\pm0.8}$        & 78.0$^{\pm1.4}$        & 73.7$^{\pm1.8}$        \\
                                & E2E                            & 42.7$^{\pm2.5}$        & 54.0$^{\pm1.2}$        & 49.0$^{\pm0.6}$        & 49.3$^{\pm1.5}$        & 43.5$^{\pm1.3}$        & 22.7$^{\pm5.2}$        & 58.0$^{\pm1.5}$        & 53.7$^{\pm1.1}$        & 55.0$^{\pm3.8}$        & 54.5$^{\pm0.7}$        & 63.0$^{\pm1.0}$        & 53.7$^{\pm1.2}$        \\
                                & ACTE                           & 37.4$^{\pm1.0}$        & 42.0$^{\pm2.8}$        & 43.7$^{\pm1.4}$        & 42.5$^{\pm0.9}$        & 41.5$^{\pm2.3}$        & 25.4$^{\pm2.1}$        & 60.9$^{\pm1.6}$        & 54.6$^{\pm5.0}$        & 56.9$^{\pm3.2}$        & 54.3$^{\pm1.6}$        & 49.8$^{\pm3.4}$        & 47.2$^{\pm0.7}$        \\
                                & TASD                           & 34.6$^{\pm1.0}$        & 36.6$^{\pm2.7}$        & 36.9$^{\pm3.0}$        & 36.7$^{\pm1.3}$        & 36.3$^{\pm2.0}$        & 17.3$^{\pm1.8}$        & 56.7$^{\pm0.3}$        & 50.9$^{\pm1.5}$        & 55.4$^{\pm2.0}$        & 49.5$^{\pm2.6}$        & 45.8$^{\pm1.7}$        & 47.0$^{\pm1.3}$        \\ \cdashlinelr{1-14}
\multirow{4}{*}{LLaMA~3~8B}               & ACSA                           & 65.2$^{\pm1.3}$        & 71.4$^{\pm1.3}$        & 66.8$^{\pm3.3}$        & 67.5$^{\pm3.5}$        & 71.1$^{\pm2.6}$        & 61.1$^{\pm3.0}$        & 79.7$^{\pm1.6}$        & 75.6$^{\pm2.0}$        & 79.5$^{\pm2.3}$        & 76.8$^{\pm2.8}$        & 77.8$^{\pm2.5}$        & 77.3$^{\pm1.1}$        \\
                                & E2E                            & 37.9$^{\pm3.0}$        & 47.2$^{\pm5.0}$        & 41.7$^{\pm3.0}$        & 43.3$^{\pm3.5}$        & 53.1$^{\pm2.2}$        & 26.6$^{\pm9.7}$        & 63.3$^{\pm5.8}$        & 60.6$^{\pm3.6}$        & 61.5$^{\pm3.7}$        & 55.5$^{\pm1.6}$        & 64.2$^{\pm2.8}$        & 62.0$^{\pm1.7}$        \\
                                & ACTE                           & 45.3$^{\pm4.7}$        & 44.8$^{\pm8.5}$        & 38.1$^{\pm3.5}$        & 40.0$^{\pm6.0}$        & 50.6$^{\pm4.1}$        & 29.9$^{\pm7.1}$        & 60.1$^{\pm3.2}$        & 54.5$^{\pm2.7}$        & 51.0$^{\pm4.2}$        & 50.7$^{\pm1.0}$        & 48.5$^{\pm4.6}$        & 51.3$^{\pm2.8}$        \\
                                & TASD                           & 33.3$^{\pm2.9}$        & 39.5$^{\pm7.5}$        & 31.5$^{\pm3.2}$        & 29.8$^{\pm5.7}$        & 46.5$^{\pm2.6}$        & 22.8$^{\pm7.2}$        & 55.8$^{\pm3.6}$        & 54.0$^{\pm3.1}$        & 49.5$^{\pm3.9}$        & 47.1$^{\pm0.7}$        & 42.1$^{\pm6.5}$        & 47.6$^{\pm3.5}$        \\ \cdashlinelr{1-14}
\multirow{4}{*}{Orca~2~13B}               & ACSA                           & \textbf{70.6}$^{\pm0.1}$        & \textbf{74.8}$^{\pm0.7}$        & \textbf{73.8}$^{\pm0.5}$        & 70.8$^{\pm0.4}$        & \textbf{75.6}$^{\pm1.6}$        & \textbf{62.9}$^{\pm2.1}$        & 79.6$^{\pm0.2}$        & \textbf{81.4}$^{\pm0.3}$        & \textbf{83.8}$^{\pm0.7}$        & \textbf{81.0}$^{\pm0.8}$        & \textbf{82.3}$^{\pm0.4}$        & \textbf{80.5}$^{\pm0.2}$        \\
                                & E2E                            & \textbf{57.6}$^{\pm0.4}$        & \textbf{69.5}$^{\pm0.6}$        & \textbf{65.6}$^{\pm0.7}$        & \textbf{61.2}$^{\pm0.3}$        & \textbf{58.9}$^{\pm1.7}$        & \textbf{33.8}$^{\pm0.8}$        & \textbf{76.5}$^{\pm0.6}$        & 72.0$^{\pm0.2}$        & \textbf{75.6}$^{\pm0.3}$        & \textbf{66.3}$^{\pm0.2}$        & \textbf{79.0}$^{\pm0.6}$        & \textbf{67.5}$^{\pm1.1}$        \\
                                & ACTE                           & \textbf{54.6}$^{\pm0.4}$        & \textbf{62.6}$^{\pm1.2}$        & 57.1$^{\pm2.4}$        & \textbf{53.4}$^{\pm3.8}$        & \textbf{59.3}$^{\pm1.1}$        & \textbf{38.0}$^{\pm2.2}$        & \textbf{73.8}$^{\pm0.2}$        & \textbf{68.9}$^{\pm0.1}$        & \textbf{71.8}$^{\pm0.1}$        & \textbf{63.9}$^{\pm0.7}$        & \textbf{73.1}$^{\pm0.4}$        & \textbf{63.3}$^{\pm0.2}$        \\
                                & TASD                           & \textbf{49.7}$^{\pm0.5}$        & \textbf{58.0}$^{\pm1.1}$        & \textbf{56.1}$^{\pm1.0}$        & \textbf{50.2}$^{\pm1.1}$        & \textbf{55.6}$^{\pm1.4}$        & \textbf{31.4}$^{\pm1.6}$        & \textbf{71.7}$^{\pm0.2}$        & \textbf{67.2}$^{\pm0.3}$        & \textbf{68.3}$^{\pm1.6}$        & \textbf{59.1}$^{\pm1.4}$        & \textbf{69.2}$^{\pm0.6}$        & \textbf{60.4}$^{\pm0.1}$        \\ \cdashlinelr{1-14}
\multirow{4}{*}{Orca~2~7B}               & ACSA                           & 70.4$^{\pm0.7}$        & 74.6$^{\pm0.5}$        & 71.8$^{\pm1.5}$        & 67.7$^{\pm0.9}$        & 75.0$^{\pm0.8}$        & 59.9$^{\pm1.3}$        & \textbf{81.6}$^{\pm0.9}$        & 79.9$^{\pm0.5}$        & 81.8$^{\pm0.5}$        & 80.0$^{\pm1.1}$        & 81.3$^{\pm1.2}$        & 79.0$^{\pm0.2}$        \\
                                & E2E                            & 55.2$^{\pm1.4}$        & 69.2$^{\pm1.3}$        & 63.4$^{\pm1.2}$        & 60.6$^{\pm1.0}$        & 54.0$^{\pm1.7}$        & 29.3$^{\pm0.9}$        & 76.2$^{\pm0.3}$        & \textbf{75.0}$^{\pm1.0}$        & 73.5$^{\pm0.2}$        & 59.7$^{\pm6.4}$        & 77.6$^{\pm0.9}$        & 64.4$^{\pm1.1}$        \\
                                & ACTE                           & 54.2$^{\pm0.2}$        & 61.1$^{\pm1.2}$        & \textbf{57.4}$^{\pm1.7}$        & 52.6$^{\pm0.3}$        & 55.3$^{\pm0.7}$        & 29.8$^{\pm5.5}$        & 72.8$^{\pm0.4}$        & 68.7$^{\pm1.0}$        & 71.0$^{\pm1.1}$        & 59.2$^{\pm0.4}$        & 67.8$^{\pm0.3}$        & 62.6$^{\pm1.1}$        \\
                                & TASD                           & 45.1$^{\pm1.0}$        & 55.9$^{\pm0.6}$        & 49.5$^{\pm1.1}$        & 45.2$^{\pm1.2}$        & 47.6$^{\pm5.1}$        & 24.8$^{\pm1.8}$        & 70.0$^{\pm0.3}$        & 66.6$^{\pm1.0}$        & 66.1$^{\pm3.4}$        & 56.0$^{\pm1.1}$        & 68.0$^{\pm0.1}$        & 60.2$^{\pm0.3}$        \\ \bottomrule
\end{tabular}
\end{adjustbox}
\label{tab:res_llm_cl}
\end{table}

\subsection{Inference and Training Speed}
Table~\ref{tab:speed} presents the average absolute and relative times for training one epoch and inference time per example for various models on the TASD task, with English as the source language and Czech as the target language. The absolute times are measured in seconds, while the relative times indicate a comparison against the baseline model, mT5. All experiments were performed on a same hardware for comparability.

\begin{table}[ht!]
    \centering
    \caption{Average absolute and relative training time per epoch and inference time per example for different models on the TASD task, with English as the source language and Czech as the target language.}
    \begin{adjustbox}{width=0.7\linewidth}
        \begin{tabular}{@{}lrrrr@{}}
            \toprule
            \multirow{2}{*}{\textbf{Model}} & \multicolumn{2}{l}{\textbf{Training Time Per Epoch}} & \multicolumn{2}{l}{\textbf{Inference Time Per Example}} \\ \cmidrule(lr){2-3}  \cmidrule(lr){4-5}
                                            & Absolute [s]                  & Relative                 & Absolute [s]                   & Relative                   \\ \midrule
            mT5                             & 178                           & 1.00                     & 0.28                           & 1.00                       \\
            Multi-tasking mT5               & 1,326                          & 7.45                     & 0.27                           & 0.96                       \\
            Orca 2 13B                      & 1,674                          & 9.40                     & 4.29                           & 15.32                      \\
            \bottomrule
        \end{tabular}
    \end{adjustbox}
    \label{tab:speed}
\end{table}

The mT5 model serves as the reference, with a relative training time of 1.00. The multi-tasking variant of mT5, while more computationally demanding during training (7.45 times slower than mT5), exhibits a similar inference time, as inference follows the same procedure as the baseline mT5 model. The Orca~2~13B model is significantly slower, requiring 9.40 times the training time of mT5 and a much higher inference time, 15.32 times that of mT5. This indicates that while larger models like Orca~2~13B may offer performance gains, they come with substantial computational costs during both training and inference.

\subsection{Error Analysis}
\label{subsec:error}
\begin{figure}[ht!]
    \centering
    \begin{adjustbox}{width=0.45\linewidth}
        \begin{tikzpicture}
            \begin{axis}[
                ybar,
                bar width=9pt,
                xtick={0,1,2},
                ytick={0,10,20,30,40,50,60},
                xticklabels={\scriptsize aspect term,\scriptsize category,\scriptsize polarity},
                ymin=0,
                ymax=65,
                xmin=-0.5,
                xmax=2.5,
                ymajorgrids=true,
                ylabel={\footnotesize Number of errors},
                xlabel={\footnotesize Sentiment element},
                legend style={at={(1,1)}, anchor=north east, font=\footnotesize},
                ]
                \addplot[black,fill=lightblue,postaction={pattern=north west lines}] coordinates {(0,49) (1,37) (2,8)};
                \addplot[black,fill=lightorange,postaction={pattern=north east lines}] coordinates {(0,39) (1,34) (2,9)};
                \addplot[black,fill=lightgreen,postaction={pattern=grid}] coordinates {(0,55) (1,34) (2,13)};
                \legend{Orca~2~13B,mT5 w/ CD, mT5 w/o CD}
            \end{axis}
        \end{tikzpicture}
    \end{adjustbox}
    \caption{Number of error types for Orca~2~13B and mT5 with and without constrained decoding (CD) on the Spanish target language and the TASD task.}
    \label{fig:error}
\end{figure}
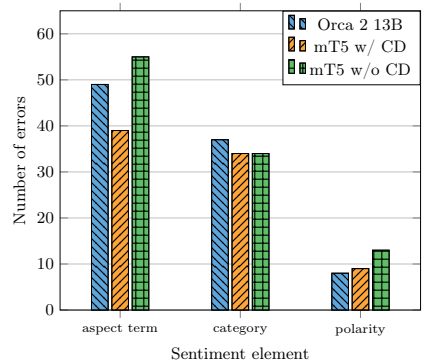

To gain insights into the challenges of sentiment prediction, we conduct an error analysis focusing on identifying the most difficult sentiment elements to predict. We manually investigate 100 random test samples with predictions from the best-performing run of the mT5, both with and without constrained decoding, for a few language combinations (cs$\rightarrow$en, en$\rightarrow$cs, en$\rightarrow$es, en$\rightarrow$nl, en$\rightarrow$fr) for the TASD task. Figure~\ref{fig:error} shows the analysis for the en$\rightarrow$es combination, alongside results with Orca~2~13B.

\subsubsection{Output Format}
One key challenge is producing the correct output format, which is crucial for target extraction. The models occasionally struggle with this, sometimes duplicating outputs and reducing the diversity of generated sentiment triplets. Although we ensure that identical triplets are not counted multiple times (thus not impacting the results), this repetition limits the models from generating unique outputs. It potentially causes them to miss specific prediction targets.

\subsubsection{Aspect Term Prediction}
The primary source of error lies in aspect term prediction. As mentioned in the results for each task, the model sometimes generates the aspect term in the source language instead of the target language, a problem mitigated by constrained decoding. Constrained decoding helps mitigate this problem by restricting the generated tokens to only those in the input target language sentence, significantly reducing the available tokens pool.

Constrained decoding also helps reduce other errors, such as correcting typos and inventing words. For example, if the review contains the typo \textit{\quotes{se\textbf{vr}ice}}, the model might generate the corrected word \textit{\quotes{service}}. Without constrained decoding, the model also sometimes invents words. For instance, some reviews contain implicit opinions about the ambience, leading the model to generate \textit{\quotes{ambience}} instead of \textit{\quotes{it}} (an implicit aspect term). Constrained decoding reduces the generation of text that is not present in the original review or in a modified format.

Additionally, the models frequently generate partial aspect terms instead of complete ones, such as \textit{\quotes{steak}} instead of \textit{\quotes{Rump steak}}. Furthermore, the models may blend parts of the review, leading to outputs that do not match the original text's specific form. For instance, instead of \textit{\quotes{Green Curry Ramen}}, the model might generate \textit{\quotes{Green Ramen}}, a phrase not in the original review. Furthermore, the models occasionally produce lowercase output even when the original text contains uppercase letters.

\subsubsection{Aspect Categories}

In terms of aspect categories, errors are less frequent than in the case of aspect terms. The models frequently omit the less common categories, such as \textit{\quotes{location general}} or \textit{\quotes{drinks style\_options}}. The models often confuse the \textit{\quotes{restaurant miscellaneous}} and \textit{\quotes{restaurant general}} classes, which are often inconsistent in the annotations. Some categories occur only in one or a few languages; for instance, \textit{\quotes{food general}} appears solely in the Dutch test set, making it impossible for the classifier to learn from other source languages.

\subsubsection{Sentiment Polarity}

The most common error concerning sentiment polarity is in predicting the \textit{\quotes{neutral}} class, possibly due to imbalanced label distribution, since the \textit{\quotes{neutral}} class is the least frequent in all datasets.

\subsubsection{Dataset Labelling}

Additionally, we identified mistakes in the dataset labels. For example, in the test part of the English dataset, the aspect \textit{\quotes{Service}} in the sentence \textit{\quotes{Worst Service I Ever Had}} is labelled as \textit{\quotes{positive}}, despite being clearly \textit{\quotes{negative}}. Similarly, we noticed inconsistencies in the datasets, such as in the sentence \textit{\quotes{One of the best hot dogs I have ever eaten}}, where the expression \textit{\quotes{hot dogs}} is not labelled as an aspect term for the \textit{\quotes{food quality}} category; instead, it is labelled as an implicit aspect term (\textit{\quotes{NULL}}), contrary to other examples. These labelling errors could negatively impact the final scores of evaluated models.

\section{Discussion \& Recommendations}
\label{sec:discussion}

The results presented in Section~\ref{sec:results} underscore the effectiveness of constrained decoding in cross-lingual aspect-based sentiment analysis, providing a practical alternative to translation tools, which can be tricky~\cite{zhang-etal-2021-cross} or ineffective~\cite{li2020unsupervised}. Constrained decoding significantly improves sequence-to-sequence models by addressing errors in aspect term prediction. These errors include generating the aspect terms in the source language instead of the target language and predicting aspect terms not present in the original review text. On average, constrained decoding improves the results by 5\% in cross-lingual settings.

Sequence-to-sequence models offer advantages over encoder-based models used in previous cross-lingual ABSA studies due to their capability to detect implicit aspect terms and assign multiple sentiment polarities to a single aspect term, thereby providing a more comprehensive evaluation. Additionally, these models can be adapted for various ABSA tasks through straightforward changes to the output format. In contrast, encoder-based models require specialized architectures for complex ABSA tasks involving multiple sentiment elements. Moreover, sequence-to-sequence models facilitate multi-task fine-tuning, allowing simultaneous predictions for different tasks. These attributes make sequence-to-sequence models preferable for ABSA applications compared to encoder-based models. Among the evaluated models, mT5 consistently outperforms mBART.

Constrained decoding plays a crucial role in achieving competitive results with multi-tasking models compared to specialized models in aspect term prediction tasks, where it improves the results by more than 10\%. While our findings do not favour multi-tasking over single-task specialization, a consistent trend suggests that multi-tasking matches or surpass single-task results in most cases. Given the substantial additional resources required for fine-tuning multi-tasking models – particularly the sixfold increase in training examples compared to single-task models, which prolongs fine-tuning duration – alongside the absence of clear performance advantages, we recommend fine-tuning specialized models when focusing on specific ABSA tasks. However, multi-tasking models offer a viable option for applications requiring simultaneous handling of multiple ABSA tasks, performing comparably to specialized models without significant performance trade-offs.

Using large language models in zero-shot and few-shot settings yields poor results for compound ABSA tasks but provides quick results without fine-tuning. ChatGPT consistently outperforms other evaluated LLMs in zero-shot and few-shot scenarios across all tasks and languages. However, fine-tuning LLMs can perform well in monolingual settings, often outperforming the mT5 model except for some languages.

Fine-tuning LLMs in cross-lingual settings can achieve performance similar to the mT5 model with constrained decoding in some cases. Fine-tuned LLMs generally perform better than mT5 when English is the target language. Nevertheless, using English as the target language is impractical and uncommon in real-world scenarios where English is typically the source language. For the more common scenario where English is the source language, mT5 often outperforms LLMs, sometimes by more than 10\% for certain language combinations. The choice of LLM is crucial. Only Orca~2~13B achieves comparable results to mT5 among the evaluated LLMs. Since LLMs are predominantly pre-trained on English data\footnote{For example, 90\% of pre-training data for LLaMA~2 models is in English~\cite{touvron2023llama2}.} and multilingual open-source LLMs are still evolving, we recommend using mT5 with constrained decoding. Additionally, fine-tuning LLMs on consumer GPUs requires special techniques such as quantization and parameter-efficient fine-tuning, whereas fine-tuning the mT5 model does not require these techniques. Moreover, the training and inference times of LLMs are significantly higher compared to the mT5 model, which could be a considerable drawback in scenarios where rapid processing is crucial.

The error analysis presented in Section~\ref{subsec:error} reveals recurring dataset issues, including inconsistent labelling of aspect terms and categories and mislabelled sentiment polarity. These findings highlight the pivotal role of high-quality, consistent datasets in the training and evaluation of models. Addressing these challenges through enhanced data curation and rigorous cleaning processes can significantly improve model performance.

Based on our findings, we provide recommendations for various scenarios in Table~\ref{tab:recommendations} to enhance cross-lingual and monolingual ABSA.
\begin{table}[ht!]
    \centering
    \caption{Recommendations for cross-lingual and monolingual ABSA.}
    \begin{adjustbox}{width=\linewidth}
    
    \begin{tabular}{@{}m{5.5cm} m{10.5cm}@{}}
    \toprule
    \textbf{Scenario} & \textbf{Recommendation} \\ \midrule
    Quick results regardless of performance & Use LLMs with zero-shot or preferably few-shot prompts; larger models generally perform better. \\ \cdashlinelr{1-2}
    High-performance monolingual results & Fine-tune LLMs, mT5 or a monolingual sequence-to-sequence model if available. \\ \cdashlinelr{1-2}
    High-performance cross-lingual results & Use mT5 with constrained decoding or fine-tune the Orca~2~13B model. \\ \cdashlinelr{1-2}
    High-performance cross-lingual results for multiple tasks simultaneously & Use multi-tasking mT5 with constrained decoding. \\ \cdashlinelr{1-2}
    Quick training and inference & Fine-tune mT5 or a monolingual sequence-to-sequence model if available. \\ 
    \bottomrule
    \end{tabular}
    \end{adjustbox}
\label{tab:recommendations}
\end{table}

\section{Conclusion}
\label{sec:conclusion}
This paper presents a comprehensive study of cross-lingual and monolingual aspect-based sentiment analysis, with a primary focus on cross-lingual analysis, leveraging sequence-to-sequence models and large language models. We conduct extensive experiments across seven languages and six different ABSA tasks using restaurant domain datasets. Notably, we are the first to access four out of these six ABSA tasks evaluated. Our pioneering use of sequence-to-sequence models for cross-lingual ABSA, alongside the first application of fine-tuned LLMs in this domain, expands the frontier of cross-lingual ABSA research.

Our approach, centred on constrained decoding combined with sequence-to-sequence models, yields significant improvements in cross-lingual ABSA performance while eliminating reliance on external translation tools. Specifically, using constrained decoding improves the target-aspect-sentiment detection task results by 5\% compared to not using it, demonstrating its effectiveness in enhancing model accuracy across diverse languages.

We surpass previous state-of-the-art results in cross-lingual ABSA tasks, showcasing the effectiveness of our proposed methodology. Moreover, our method supports multi-tasking capabilities, enabling simultaneous resolution of multiple ABSA tasks. Constrained decoding enhances the performance of multi-tasking models by more than 10\%, underscoring its efficacy in addressing complex linguistic tasks across diverse languages.

In addition to evaluating LLMs in zero-shot, few-shot, and fine-tuning scenarios in both monolingual and cross-lingual settings, we demonstrate that while LLMs struggle in zero-shot and few-shot contexts compared to smaller fine-tuned models, fine-tuning boosts their performance significantly in monolingual tasks. However, LLMs generally lag behind smaller models equipped with constrained decoding in cross-lingual settings.

Furthermore, we provide detailed error analysis to highlight the primary challenges in cross-lingual ABSA, and we compare different models in terms of training and inference speed. Based on our extensive experimentation and evaluation, we propose a comprehensive set of recommendations tailored for various real-world scenarios in monolingual and cross-lingual ABSA applications. These recommendations aim to guide practitioners and researchers towards effective model selection and deployment strategies in diverse linguistic contexts.

For future research, potential directions could include verifying the effectiveness of our cross-lingual ABSA methods across different domains and languages, expanding beyond the restaurant domain. These efforts would involve creating and utilizing datasets  designed explicitly for cross-lingual aspect-based sentiment analysis in various contexts. Currently, the availability of datasets in different languages for various domains is limited. Additionally, exploring cross-domain, cross-lingual ABSA could simultaneously assess the potential for transferring sentiment-related knowledge across different domains and languages. This challenging task could lead to significant advancements in cross-lingual sentiment analysis with practical applications across diverse fields.

\begin{credits}
\subsubsection{\ackname} This work has been supported by the Grant No. SGS-2025-022 -- New Data Processing Methods in Current Areas of Computer Science.
% The work of the other authors has been supported by the project R\&D of Technologies for Advanced Digitalization in the Pilsen Metropolitan Area (DigiTech) No. CZ.02.01.01/00/23\_021/0008436. 
Computational resources were provided by the e-INFRA CZ project (ID:90254), supported by the Ministry of Education, Youth and Sports of the Czech Republic.

\subsubsection{\discintname}
The authors have no competing interests to declare that are relevant to the content of this article.
\end{credits}

\bibliographystyle{splncs04}
\bibliography{bibliography}

\end{document}